\PassOptionsToPackage{prologue,dvipsnames}{xcolor}
\documentclass{article} 
\usepackage{iclr2025_conference,times}

\usepackage{amsmath,amsfonts,bm}

\def\eqref#1{equation~\ref{#1}}

\def\1{\bm{1}}

\DeclareMathAlphabet{\mathsfit}{\encodingdefault}{\sfdefault}{m}{sl}
\SetMathAlphabet{\mathsfit}{bold}{\encodingdefault}{\sfdefault}{bx}{n}

\usepackage{hyperref}
\usepackage{url}

\usepackage{amsmath}
\usepackage{amsfonts} 
\usepackage{amssymb}
\usepackage{multirow}
\usepackage{graphicx}
\usepackage{tabularx}
\usepackage{adjustbox}
\usepackage{pifont}
\usepackage{booktabs}
\usepackage{subfig}

\usepackage{colortbl}
\definecolor{Orange}{rgb}{0.9,0.5,0}
\definecolor{NavyBlue}{rgb}{0.1, 0.4, 0.8}
\definecolor{Magenta}{rgb}{0.8, 0.1, 0.6}
\definecolor{F7E0D5}{RGB}{247,224,213}
\definecolor{Gray}{gray}{0.9}
\colorlet{Light}{white!0!F7E0D5}

\usepackage{xspace}
\newcommand{\ours}{COGT\xspace}
\newcommand{\cmark}{\ding{51}}%
\newcommand{\xmark}{\ding{55}}%

\usepackage[capitalise]{cleveref}  
\crefname{section}{sec.}{sec.}
\crefname{equation}{eq.}{eq.}
\crefname{figure}{fig.}{fig.}
\crefname{table}{tab.}{tab.}
\crefname{appendix}{app.}{app.}

\title{Causal Graphical Models for Vision-Language Compositional Understanding}

\author{Fiorenzo Parascandolo, Nicholas Moratelli, Enver Sangineto, Lorenzo Baraldi \& Rita Cucchiara \\
AImageLab, Department of Engineering ``Enzo Ferrari''\\
University of Modena and Reggio Emilia\\
\texttt{name.surname@unimore.it} \\
}

\iclrfinalcopy 
\begin{document}

\maketitle
\begin{abstract}
Recent work has empirically shown that Vision-Language Models (VLMs) struggle to fully understand the compositional properties of the human language, usually modeling an image caption as a ``bag of words''. As a result, they perform poorly on compositional tasks, which require a deeper understanding of the different entities of a sentence (subject, verb, etc.) jointly with their mutual relationships in order to be solved. In this paper, we model  the dependency relations among textual and visual tokens using a {\em Causal Graphical Model} (CGM), built using a {\em dependency parser}, and we train a decoder conditioned by the VLM visual encoder. 
Differently from standard autoregressive or parallel predictions,
our decoder's generative process is partially-ordered following the CGM structure. This structure encourages the decoder to learn only the main causal dependencies in a sentence discarding spurious correlations.
Using extensive experiments on five compositional benchmarks, we show that our method significantly outperforms all the state-of-the-art compositional approaches by a large margin, and it also improves over  methods trained  using much larger datasets.
Our model weights and code are publicly available.\footnote{\url{https://aimagelab.github.io/COGT}}
\end{abstract}

\section{Introduction}
\label{sec.Introduction}

Vision-Language  Models (VLMs) have shown impressive results in  different tasks such as, for instance, zero-shot classification, image-text retrieval, 
vision-question answering, 
image-captioning, and many others \citep{radford2021learning,BLIP-2,FLAVA,LLaVA}.
However, despite this success, most VLMs still struggle in understanding the compositional nature of the human language.
For instance, \citet{NegCLIP}  empirically showed that common VLMs usually do not consider the order and the syntactic/semantic relations of words in a sentence, which is treated as a {\em bag of words}, where ``the horse is eating the grass'' and ``the grass is eating the horse'' can easily be confused.
Jointly with \citet{NegCLIP}, many other authors have recently proposed different {\em compositional benchmarks} which confirm the poor performance of common VLMs when tested against compositional tasks \citep{sugarcrepe,vl_checklist,colorswap,clip-spec}.
One of the probable reasons of this bag-of-words behavior is the 
contrastive loss used in CLIP \citep{radford2021learning} (and in other VLMs), which compares a single vector representing the
textual encoder's output with a single vector representing the
visual encoder's output, sacrificing textual and visual  details \citep{NegCLIP,kamath2023text,Distilling}.
Another reason is the
low quality of
the  captions  
used for 
VLM pre-training, which are usually noisy or do not describe the details of the image and the interactions among its objects \citep{DAC}.

Most of the compositional methods that have  been recently proposed to alleviate this problem focus on creating 
annotations with a richer compositional structure, used to fine-tune a VLM  \citep{NegCLIP,DAC,cascantebonilla2023goingnounsvision}.
For instance, NegCLIP \citep{NegCLIP} creates
{\em hard negatives}, in which the original caption is modified swapping the positions of some words, and these hard negatives are used jointly with common negatives to fine-tune CLIP  using the standard contrastive loss. However, the automatic creation of hard negatives is itself noisy, leading to captions which often do not have  a correct syntactic/semantic meaning (this problem is inherited by some compositional benchmarks, see~\Cref{sec.Experiments}). In \citep{Cap}, a VLM 
is pre-trained from scratch using a captioning strategy and a huge private dataset. Specifically, the authors propose both Cap, where the pre-training strategy is a standard {\em AutoRegressive} (AR) next-token prediction, and CapPa, where the AR training is mixed with a {\em parallel} training \citep{bolelli2018hierarchical}, in which all the textual tokens are simultaneously predicted. \citet{Cap} show that both Cap and CapPa achieve excellent results on compositional tasks, and argue that 
a {\em generative training} encourages the VLM to focus on  fine-grained descriptions of the visual content.

\begin{figure*}[t]
    \centering
    \includegraphics[width=0.6\linewidth]{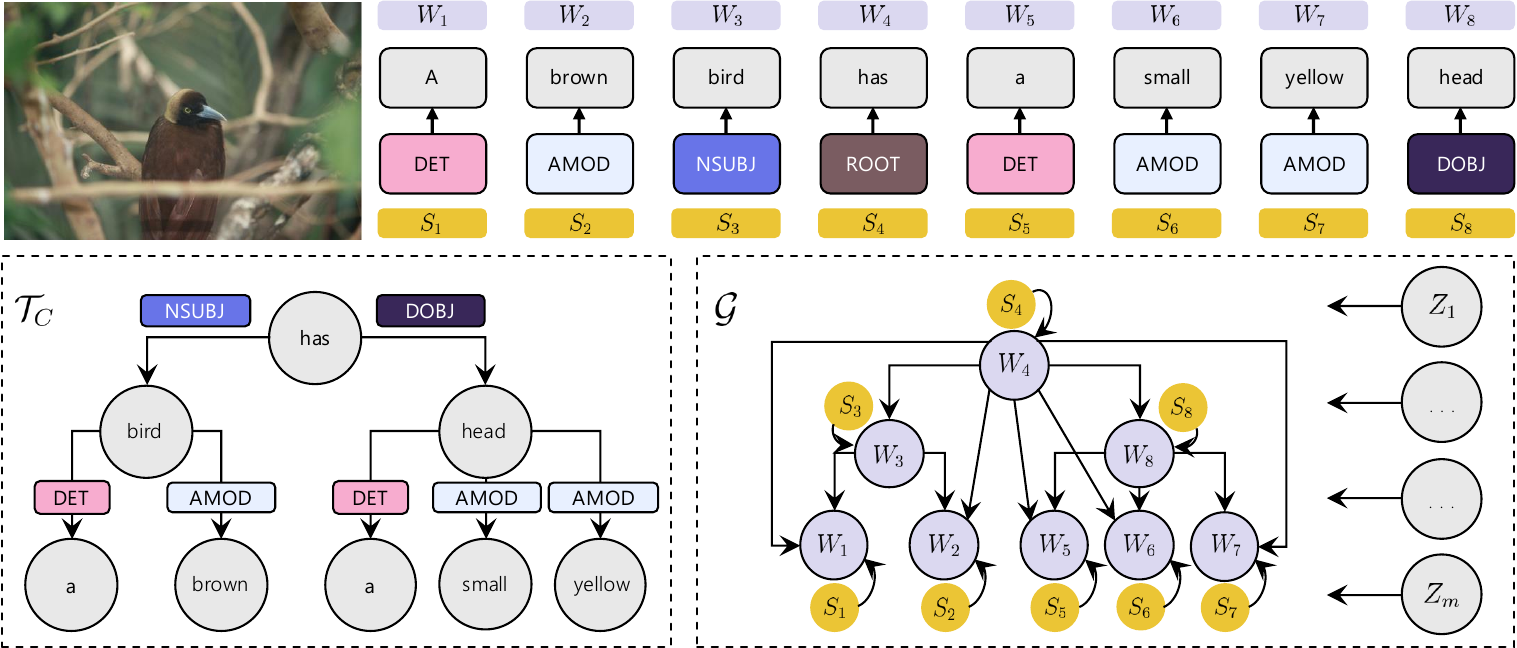}
    \caption{Dependency relations between words in a sentence. On the left, the DT ($\mathcal{T}$) extracted from the caption shown above using a dependency parser \citep{OurDependencyParser}. 
    On the right, the corresponding CGM ($\mathcal{G}$). 
    To improve readability, in $\mathcal{G}$ we use different colors for  different variable types and we omit 
    the causal dependencies between visual ($Z$) and textual ($W$) variables.}
    \label{fig.parser}
\end{figure*}

In this paper, inspired by Cap and CapPa, we propose a VLM adaptation approach for compositional reasoning which is based on a decoder trained with a captioning strategy. 
However, differently from 
the standard fully-sequential AR and the parallel predictions used in~\citep{Cap},  we propose a  partially ordered, semi-parallel AR prediction strategy which is guided by the dependency relations of a {\em Causal Graphical Model} (CGM) \citep{TowardCausal}. In more detail, we use  an off-the-shelf  {\em dependency parser} \citep{OurDependencyParser},
which  
creates a syntactic  tree from a given textual sentence. Specifically, given a caption, a dependency parser 
 automatically builds a  {\em Dependency Tree} (DT), in which each node is associated with a caption word and each edge represents a syntactic dependency relation between two words (\Cref{fig.parser}). The DT, jointly with the visual features extracted from the image using a frozen  visual encoder, are used to build a CGM, which describes the dependency relations among image patches and textual tokens. Our token prediction strategy is based on the dependency relations contained in this CGM. The rationale behind this approach is illustrated in \Cref{fig.parser} using the caption ``A brown bird has a small
yellow head''. For instance, in the resulting DT, the adjective ``brown'' depends on the noun ``bird''. 
However, using a standard AR approach, where the token prediction order follows the English grammar, the captioning model should predict ``brown'' before knowing that this adjective refers to  ``bird'', which is a quite ambiguous task, since many objects may be brown in the image.
Conversely, when our model predicts the adjective (``brown''), it knows the noun (``bird'') it refers to, thus the word generation can be specific to the entities, the attributes and the relations contained  in the input image. 
Generally speaking, we factorize the joint distribution of all the caption words following the {\em disentangled  factorization} of a CGM \citep{TowardCausal}, and our semi-parallel AR model predicts a token conditioned only on the  tokens on which it depends.  For instance, in the example of \Cref{fig.parser}, ``small'' and ``yellow'' are predicted in parallel and they are conditionally independent given ``head'', thus no statistical dependence is learned between these two words. The advantage of this strategy is that the decoder can focus on learning only the main causal dependency relations, ignoring possible spurious associations 
\citep{PEARL1995789}
induced by the sequential  order of the words in a natural language sentence. 
Moreover, we use the same prediction strategy also at inference time, when we compute the likelihood of a candidate caption. In this case too, the use of the CGM makes 
 the likelihood estimation
 independent of spurious associations due to the sequential order of the words.

We validate our method using  
different  VLMs (CLIP,  XVLM \citep{xvlm} and InstructBLIP \citep{instructblip}).
Using extensive experiments with five compositional datasets, we show that our approach largely outperforms
all previous works, setting a new {\em state of the art} in all the evaluated benchmarks,
and that it also improves on Cap and CapPa, despite being trained on much less data.

\section{Related work}
\label{sec.RelatedWork}

{\bf Compositional Methods.} Most  of the compositional methods  are based on 
creating {\em annotated training samples} 
which force the VLM to acquire compositional knowledge.
 For instance, \citep{NegCLIP,zhang2024contrastingintramodalrankingcrossmodal,huang2023structureclipscenegraphknowledge,buettner2023investigatingroleattributecontext,momeni2023verbsactionimprovingverb,Doveh_2023_CVPR,singh-etal-2023-coarse,oh2024preserving,3VL,Incorporating} 
 use either a rule-based method or a Large Language Model (LLM) to
 create {\em hard negatives} (\Cref{sec.Introduction}), which typically consist in 
  replacing or swapping the position of some words in the ground-truth caption associated with a training image. In \citep{cascantebonilla2023goingnounsvision}, dense captions are constructed using 
synthetic videos created with a 
 3D physics-based simulator, while 
 \citep{s2024figclipfinegrainedclipadaptation} use real videos.
In DAC \citep{DAC}, dense captions are created by combining the results of either an LLM (GPT-NEO-2.7B) or a segmentation network (SAM \citep{kirillov2023segany}) with a captioner (BLIP-2 \citep{BLIP-2}). SAM is also used 
in \citep{sahin2023enhancingmultimodalcompositionalreasoning}
jointly with Stable Diffusion \citep{stableDiffusion}
to generate hard negative {\em images}. 
Moreover, Stable Diffusion is  used in \citep{li2023diffusionmodelsecretlyzeroshot,clark2023texttoimagediffusionmodelszeroshot,krojer2023diffusionmodelsvisionandlanguagereasoners} 
as an alternative VLM. The main idea is that the noise prediction error of the Diffusion Model (DM) \citep{DDPMs}, obtained
by feeding Stable Diffusion with a corrupted version of the test image and a given caption, can be used as an estimate of the image-caption similarity.
Finally, Stable Diffusion is used in \citep{Distilling} as an additional regularization loss to fine-tune CLIP.

Apart from DM-based methods, most of the compositional approaches are based on fine-tuning or adapting CLIP. For instance, 
\citet{zhang2024contrastingintramodalrankingcrossmodal} use an hinge loss with a curriculum-learning based adaptive margin, 
while \citet{DAC}
use a Multiple Instance
Learning  loss.
Curriculum learning
is used also in \citep{singh-etal-2023-coarse}, while
\citet{zheng2024iteratedlearningimprovescompositionality} iteratively retrain CLIP and represent an image using a sparse combination of codebook codes.
\citet{oh2024preserving} propose a local hard negative loss to fine-tune CLIP which is based on a dense alignment betweeen patch embeddings and textual token embeddings.
\citet{wazni-etal-2024-verbclip} use a dependency parser (see below) to extract triplets (subject, verb, object) from a caption. Subjects and objects are represented as embedding vectors using the CLIP textual encoder, while verbs are represented by matrices that are multiplied with either the subject or the verb to change their meaning.
In \citep{li2024covlm}, CLIP is embedded in a larger VLM, which includes a detection network and an LLM. The LLM comunicates with the detection network using special tokens. 
A few methods use {\em generative pre-trained} VLMs, and their results usually show a large improvement with respect to encoder-based VLMs when applied to compositional tasks, most likely because the next-token prediction pre-training 
encourages the VLM to learn the natural language compositional characteristics. 
For instance, \citet{Incorporating} use ``Adaptive Scene Graph Tokens'' to adapt both CLIP and BLIP-2 \citep{BLIP-2}
to predict scene
graph information, and they show that the BLIP-2 based results are much higher than those based on CLIP.
\citet{CRG} use LLaVA \citep{LLaVA} and a 
classifier-free guidance strategy, in which they compare the VLM prediction on two images: the original test image and a modified version where the main objects are masked-out. BLIP is used also by \citep{Revisiting}, who focus on mitigating the linguistic bias on the VLM pre-training dataset.
 Finally, \citet{Cap} propose two VLMs, called Cap and CapPa (see \Cref{sec.Introduction}), both trained generatively.  Cap is a standard AR captioner, while CapPa is trained using a combination of $25\%$ AR next-token prediction and $75\%$ fully-parallel token prediction. 
 Inspired by the success of generative pre-training, in  this paper we propose a decoder trained using a semi-parallel prediction strategy, where the order in which future tokens are predicted depends on a CGM, and we show that it can be applied to both encoder-only and generative pre-trained VLMs, significatively boosting the results of both.

{\bf  Causal Graphical Models.} 
CGMs are used in causal learning to represent the  causal relations among a set of variables \citep{TowardCausal,Causaldiscovery}.  These  relations are supposed to be known and 
are represented  by
the edges connecting each variable (node) in the graph with the variables on which it depends (``parents''). 
The joint distribution of all the variables of the CGM is computed using the {\em disentangled factorization} 
~\citep{TowardCausal,Causaldiscovery}, given by the 
product of all the conditional distributions of each variable with respect to its parents (\Cref{app.Factorization}).
The main advantage of this factorization is that, since it is assumed to be {\em causally sufficient}, the model does not need to learn other inter-variable conditional distributions, in this way reducing the number of training samples necessary to learn the joint distribution \citep{TowardCausal}. 
As far as we know, the only work using CGMs for vision-language  compositional tasks is \citep{ComCLIP},
where 
{\em Independent Causal Mechanisms (ICMs)} \citep{DBLP:journals/corr/abs-1712-00961,goyal2021recurrent,TowardCausal}
describe the relations between the subject, the object and the action of an image. However, the method proposed in \citep{ComCLIP} is radically different from our proposal, being each 
ICM simply computed as the CLIP similarity between a word and a sub-image.

{\bf Syntactic Trees.}
In {\em Dependency Grammars}, dependency relations are  syntactic and semantic connections between words in a sentence, where one word (called ``head'') governs or determines the grammatical behavior of the ``dependent'' word \citep{Nivre2005DependencyGA}. 
Given a sentence, these dependencies
are organized in a Dependency Tree (DT), which can be automatically extracted using a parser \citep{honnibal2020spacy,zhang-etal-2020-efficient,OurDependencyParser}. 
In \citep{yang2022dependencybasedmixturelanguagemodels} a Language Model (LM) is trained to 
predict whether  the future tokens in the sequence are 
 head or  dependent of a previously observed token, and this prediction replaces the standard Maximum Likelihood Estimation objective. Similarly, in
 \citep{Deguchi2019DependencyBasedSF} the dependency relations extracted by an external parser are learned by the LM and used to modulate the Transformer 
 \citep{attention-is-all-you-need}
 attention maps.
A DT can also be used to compute a {\em syntactic distance} between words, which in turn can be used, e.g., as an additional loss \citep{du2020exploitingsyntacticstructurebetter} or to modulate 
the attention maps \citep{hou2022syntaxguidedlocalizedselfattentionconstituency}.
In the  vision-language domain, DTs are used in \citep{song2022clipmodelsfewshotlearners} to convert a textual question into a template for CLIP fine-tuning, and
in \citep{li2024learningcorrectionefficienttuning} to replace and swap words in a sentence and fine-tune a VLM using the correction of the modified sequence 
as a pretext task.
Finally,  {\em constituency parsers} group  words that belong to a specific grammatical category in 
a sub-phrase. Constituency trees are used, e.g., in \citep{3VL} to generate sub-phrase specific hard negative captions 
 or in \citep{SynCLM} to extended the contrastive loss 
 by maximizing the
similarity of  words
in the same sub-phrase.
Differently from previous work, we use a DT to extract syntactic and semantic dependencies between the words of a sentence, and we interpret these dependencies as causal relations that guide the construction of our CGM.

\section{Method}
\label{sec.Method}

Given an image-caption pair $(X,C)$, our  goal is to define a set of conditional distributions over the random variables associated with the image features and the caption words.
For this purpose, as anticipated in \Cref{sec.Introduction,sec.RelatedWork}, we use an off-the-shelf dependency parser \citep{OurDependencyParser} 
which, for a specific $C = [ w_1, ..., w_n ]$, returns a DT 
$\mathcal{T}$\footnote{Note that, for each $C$ in the training/testing set, $\mathcal{T}$  needs to be extracted only once and  can be done offline.}
(\Cref{fig.parser}), where each node corresponds to a word and each edge $(i,j)$ connects the ``dependent'' word $w_j$ with its 
``head''
$w_i$ (\Cref{sec.RelatedWork}). 
$\mathcal{T}$ contains the  syntactic and semantic dependencies between the words in $C$ \citep{Nivre2005DependencyGA}, 
and we make this dependency explicit by connecting each word to all the words it transitively depends on in the tree.
Specifically, we define a CGM  $\mathcal{G}$ by associating each word $w_j$ with a random variable $W_j$,
corresponding to a node of $\mathcal{G}$.
Moreover, we connect the  node corresponding to $W_j$ with all the variables corresponding to the ancestors of $w_j$ in $\mathcal{T}$ (\Cref{fig.parser}).

Formally, if $w_{i_1}, ..., w_{i_k}$ are the ancestors of $w_j$ in $\mathcal{T}$, then we 
assume a causal dependence between the  corresponding variables: $W_{i_1} \rightarrow W_j, ..., W_{i_k} \rightarrow W_j$.
Furthermore, the parser labels each  word in $\mathcal{T}$ with a syntactic 
 type using  a prefixed vocabulary $V$ \citep{dependency_relations,zhang-etal-2020-efficient}.
For instance, if  $type(w_j) = \texttt{nsubj} \in V$, it means that $w_j$ is a noun and it plays the role of the subject in the sentence. 
Intuitively, we can think of these  syntactic types as categorical syntactic features extracted from $C$, which we formally  describe using $n$ random variables $S_1, ..., S_n$, where each $S_j$ ranges over $V$.
In $\mathcal{G}$, we assume that each $W_j$ depends on its corresponding syntactic variable $S_j$: $S_j \rightarrow W_j$.

Finally,   we extract a set of features from $X$ using the VLM visual encoder 
${\cal E}$:
${\cal Z}  = {\cal E}(X) = \{ \pmb{z}_1, ..., \pmb{z}_m \}$
(details in \Cref{sec.Decoder}),
and, similarly to the textual case, we associate a random variable $Z_k$ to each feature $\pmb{z}_k \in {\cal Z}$.
In $\mathcal{G}$, we assume that $W_j$ depends on all the visual variables: $Z_1 \rightarrow W_j, ..., Z_m \rightarrow W_j$.
Using the above assumptions,
  we define the {\em parents} \citep{TowardCausal} of $W_j$ as:
$\mathbf{PA}(W_j) =  \{ W_{i_1}, ...,  W_{i_k}, S_j, Z_1, ..., Z_m \}$, and
 we model the conditional joint distribution of the textual variables given the visual and the syntactic variables as:

\begin{equation}
\label{eq.jointdistrib-disentangled}
    P(W_1, ..., W_n | S_1, ..., S_n, Z_1, ..., Z_m) = \prod_{j=1}^n P(W_j | \mathbf{PA}(W_j)),
\end{equation}

\noindent
where the right side of \Cref{eq.jointdistrib-disentangled} is obtained using the {\em disentangled factorization} of CGMs \citep{TowardCausal,Causaldiscovery}  and assuming that $S_1, ..., S_n$ and $Z_1, ..., Z_m$ are independent of each other
(see \Cref{app.Factorization} for more details).
In \Cref{sec.Decoder} we show how a VLM can be adapted to predict this disentangled factorization both at training and at inference time.

\begin{figure*}[t]
    \centering    \includegraphics[width=0.6\linewidth]{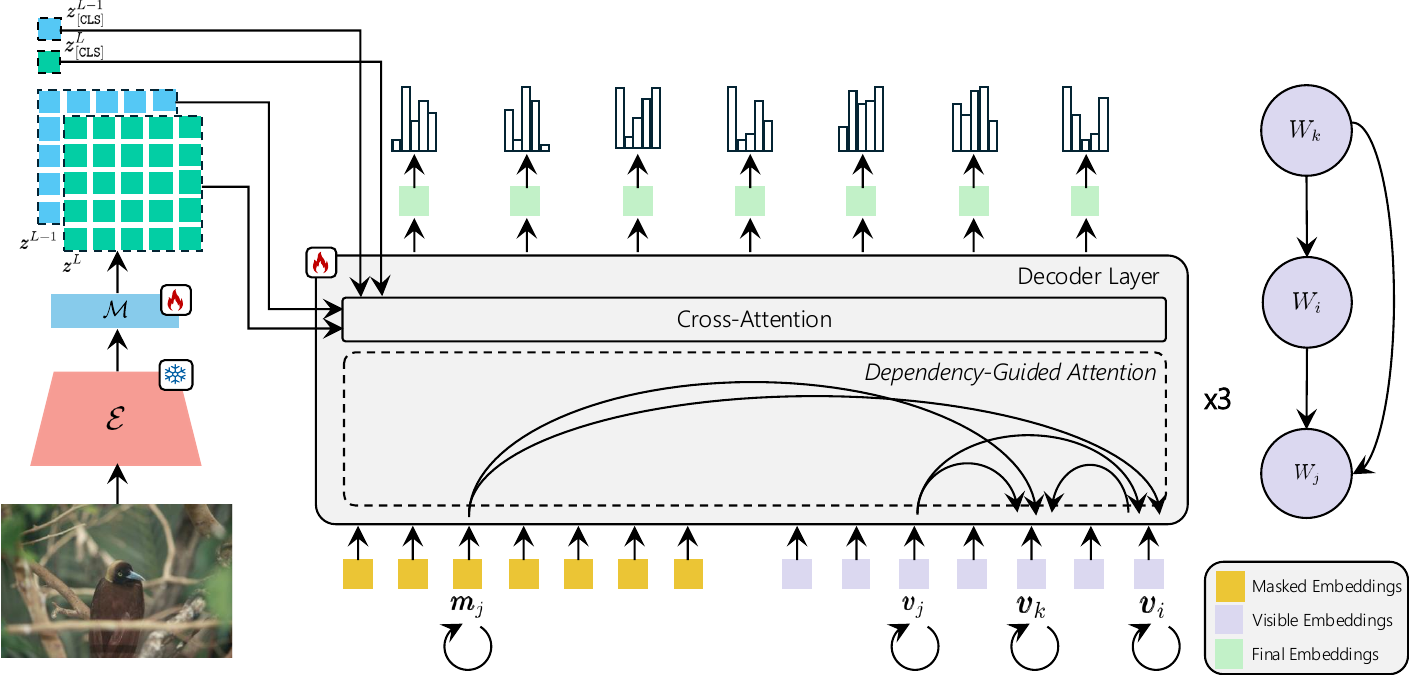}
    \caption{A schematic illustration of our decoder.}
    \label{fig.Decoder}
\end{figure*}

{\bf Discussion.}
 \citet{Cap} formulate the joint distribution of the words in $C$  using the standard AR prediction strategy commonly adopted by image captioning methods (\Cref{sec.Introduction}):

\begin{equation}
\label{eq.jointdistrib-AR}
    P(W_1, ..., W_n | Z_1, ..., Z_m) =  \prod_{j=1}^n P(W_j | W_1, ..., W_{j-1}, Z_1, ..., Z_m). 
\end{equation}

The advantage of our formulation (\Cref{eq.jointdistrib-disentangled}) over \Cref{eq.jointdistrib-AR} is that, in our case,
the model needs to learn only the inter-variable conditional distributions indicated by the dependency parser, reducing the risk of overfitting on the training data \citep{TowardCausal}. 
Specifically, the dependency parser helps in discarding those spurious {\em associations} \citep{CausalInference,PEARL1995789}
contained in  \Cref{eq.jointdistrib-AR} which depend on the sequence of words in $C$ but do not correspond to a strict semantic/syntactic relation (e.g., ``small'' and ``yellow'' in the example of  \Cref{sec.Introduction}).
In contrast, we interpret the  dependency relations extracted by a dependency parser  as {\em causal relations} because they directly model the (linguistic) influence of the ``head'' variable with respect to the generation of the ``dependent'' variable. For instance, 
the probability values of an adjective are directly influenced by the noun it refers to, because the adjective describes an attribute {\em of that noun}, thus  the corresponding conditional probability  is not a spurious association (more details in \Cref{app.Factorization}).

While the causal dependency relations in $C$ may not be exhaustively described by $\mathcal{G}$ and there may be other relations between words in $C$,
we  follow \citep{Goyal2020InductiveBF}, and we assume  that, in a symbolic domain like the natural language, 
the joint distribution over the words of a sentence should be {\em sparse}.  
This is also in line with very recent work which shows that  sparse attention in Transformers helps the network focus on the most relevant context and improves its performance removing noise \citep{selectiveattention,differentialtransformer}.
Thus, we prefer sparseness to completeness and we assume that the word dependencies extracted by a dependency parser are  causally sufficient (\Cref{sec.RelatedWork}, with more details in \Cref{app.Factorization}).
Finally, \citet{Cap} propose also a parallel prediction strategy, which corresponds to:

\begin{equation}
\label{eq.jointdistrib-parallel}
    P(W_1, ..., W_n | Z_1, ..., Z_m) = \prod_{j=1}^n P(W_j | Z_1, ..., Z_m).
\end{equation}

\noindent
In \Cref{eq.jointdistrib-parallel}, each $W_j$ is assumed to be conditionally independent from all the other textual variables given the visual variables. The empirical results reported in \citep{Cap} do not show a clear winner between the AR and the parallel prediction, and the authors use a mixture of the two strategies in training their VLM (\Cref{sec.RelatedWork}).
Conversely, in our experiments (\Cref{sec.Ablation,sec.SOTA-exp}) we show that our proposed disentangled factorization (\Cref{eq.jointdistrib-disentangled}) is a better trade-off between the conditional independence of \Cref{eq.jointdistrib-parallel} and the standard image captioning factorization of \Cref{eq.jointdistrib-AR}, and it also improves over the mixed strategy adopted in \citep{Cap}.

\subsection{Using a decoder for causal prediction}
\label{sec.Decoder}

In this section, we show how textual tokens can be generated using our CGM. 
Note that our goal is not image captioning, but  we use our method, which we call  Causally-Ordered Generative Training (\ours), for vision-language compositional  understanding.
Since CLIP is the most commonly adopted backbone by previous works on compositionality (\Cref{sec.RelatedWork}),
in the following we use CLIP as an example VLM, 
and in \Cref{sec.SOTA-exp} we  show additional results obtained  with other VLMs. 

We freeze the CLIP visual encoder (${\cal E}$) and, from a given image $X$, we extract a set of  features from the last ($L$) and the penultimate ($L-1$) layer of ${\cal E}$:
${\cal Z} = \{ \pmb{z}_{[\texttt{CLS}]}^L, \pmb{z}_1^L..., \pmb{z}_p^L, \pmb{z}_{[\texttt{CLS}]}^{L-1}, \pmb{z}_1^{L-1}..., \pmb{z}_p^{L-1} \}$. 
For the $l$-th layer of the encoder,
 $\pmb{z}_{[\texttt{CLS}]}^l$ is the embedding vector of the class token \citep{ViT}, while
 $\pmb{z}_1^l..., \pmb{z}_p^l$ are
  the embedding vectors of the patch tokens. Using a grid of $p$ patch tokens, we have $m = 2p +2$.
We use the embedding vectors of the penultimate layer jointly with the last layer features to help the model reasoning about smaller resolution objects. Indeed, previous work \citep{visual_features_penultimate,clipdinoiserteachingclipdino} showed that 
there is usually a decrease in the amount of spatial information represented in the last layer of CLIP.
Moreover, we use a  mapping network ${\cal M}$ (\Cref{fig.Decoder}) to reduce the dimensionality of the visual features 
 to match the decoder embedding size. ${\cal M}$ is composed of a linear layer, preceded and followed by LayerNorm,
 and all  features in ${\cal Z}$ are obtained as output of ${\cal M}$.
The parameters of 
 ${\cal M}$ are learned jointly with our decoder (see below) and  ${\cal M}$ is shared by 
 all  features in ${\cal Z}$ and both layers of ${\cal E}$ ($L$ and $L-1$).  

 We replace  the CLIP textual encoder with our decoder ${\cal D}$,  
 a relatively small network, composed of only three blocks   
with  $\sim$64M total parameters,
which is the  module we use to adapt CLIP to solve compositional tasks.  
\Cref{fig.Decoder} shows the architecture of ${\cal D}$, which takes as input $2 n$ tokens. The first sequence of $n$ tokens are masked tokens, while the others are visible tokens,
and we represent each $w_j$ with both a masked and a visible token.
Specifically,
to condition ${\cal D}$ with respect to the event $S_j = t$ ($t \in V$) in \Cref{eq.jointdistrib-disentangled}, we use masked tokens specific for each syntactic  type $t$ in $V$. In more detail, 
$V$ is composed of
 the 45 standard syntactic categories defined in \citep{dependency_relations}
 (see \Cref{subsec.syntactic_categories}).
We associate each category $t$ with a  masked token $\texttt{MSK}_t$.
Then, for each word $w_j \in C$,  if $type(w_j) = t$, then the masked token used for $w_j$ is  $\texttt{MSK}_t$. This is simply implemented using a lookup table of masked token embeddings, composed of 45 different initial embedding vectors (learned using standard backpropagation) and which extends the  (single) masked token used in 
common masked-token prediction tasks \citep{devlin-etal-2019-bert}.
The other $n$ tokens are visible, standard textual tokens, one for each $w_j \in C$. In this way,  $w_j$ is represented both as a visible token and as a masked token of type $t$. 
In a given layer of ${\cal D}$, these two tokens are respectively represented by the masked-token
  embedding vector $\pmb{m}_j$ and the visible-token embedding vector $\pmb{v}_j$.

  Each block of ${\cal D}$  is composed of two  layers.
  In the first layer, we compute the self-attention of each  masked embedding $\pmb{m}_j$
with itself, jointly with the attention of $\pmb{m}_j$ with all the visible embeddings $\pmb{v}_{i_1}, ...,  \pmb{v}_{i_k}$,
where 
$\mathbf{PA}(W_j) =  \{ W_{i_1}, ...,  W_{i_k}, S_j, Z_1, ..., Z_m \}$.
Note that there is no attention between $\pmb{m}_{j_1}$ and $\pmb{m}_{j_2}$, with $j_1 \neq j_2$.
In the same layer, we compute the self-attention of each visible embedding $\pmb{v}_j$
with itself, jointly with the attention of $\pmb{v}_j$ with $\pmb{v}_{i_1}, ...,  \pmb{v}_{i_k}$
(\Cref{fig.Decoder}).
Note that there is no information leak, since $\pmb{m}_j$, later used for the final prediction, has no direct or indirect access to $\pmb{v}_j$.
We call this {\em Dependency Guided Attention} to differentiate it from the standard self-attention (\Cref{fig.Decoder}).
In the second layer of each block of ${\cal D}$, 
both the masked ($\pmb{m}_j$) and the visible ($\pmb{v}_j$) embeddings 
pay attention to the visual features in ${\cal Z}$
using  cross-attention, in this way
implementing the dependence between $W_j$ and $Z_1, ..., Z_m$.
Finally, after the last block of ${\cal D}$ we discard the visible-token embeddings and we fed each masked-token final embedding to a linear  layer computing a posterior distribution over the vocabulary of textual terms. 

${\cal D}$ is trained from scratch using as the only objective the maximization of the log-likelihood of the disentangled factorization:

\begin{equation}
\label{eq.log-likelihood}
    \mathcal{L} = \log \left[ \prod_{j=1}^n P(W_j | \mathbf{PA}(W_j)) \right] = 
    \sum_{j=1}^n \log (P(W_j | \mathbf{PA}(W_j))). 
\end{equation}

{\bf Inference}
Most of the compositional tasks are modeled as 
 image-to-text retrieval tasks. In  case of \ours,
 given a testing image $X$, we compute the log-likelihood 
 of all the candidate testing captions and we select the highest scoring sentence. 
 Note that computing ${\cal Z}$ is independent of the specific caption $C$, and it can be done once for each image in the dataset.
 The log-likelihood  is computed using 
 \Cref{eq.log-likelihood}
 and a semi-parallel AR prediction strategy which follows the partial order induced by the DT. Specifically, 
 using the dependency parser we extract $\mathcal{T}$ from a candidate caption $C$. Then we 
 proceed using a {\em level order traversal} of $\mathcal{T}$, in which, starting from the root, we predict in parallel all the tokens of a given level of the tree and then we move to the next level.


\section{Experiments}
\label{sec.Experiments}

In our evaluation we use four common compositional benchmarks: ARO \citep{NegCLIP}, SugarCrepe \citep{sugarcrepe}, VL-CheckList \citep{vl_checklist} and ColorSwap \citep{colorswap}, and an additional benchmark  FG-OVD \citep{FG-OVD} which we propose in this paper.
Most of them are composed of different tasks and datasets, and we report both the task-specific and the average accuracy across all tasks.
Following \citep{zhang2024contrastingintramodalrankingcrossmodal}, 
 we do not use COCO Order and Flickr Order 
(two of the  ARO tasks)
because it has been previously showed that a ``blind'' LM, {\em with no  access to the image}, can achieve about  $99\%$  
accuracy on these tasks \citep{Cap}. The reason of this is due to the 
grammatical and semantic errors introduced in  the negatives when swapping or replacing caption words
 (see \Cref{sec.Introduction}). For instance, the sentence ``with man is wearing ears the an glasses pierced orange hat and''
 (Flickr  Order) can be easily detected as false by an LM without any visual knowledge.
In \Cref{subsec.winoground_eyes_wide_shut}
we show additional results with Winoground \citep{Winoground}, which, however, is often not used by other methods based on CLIP \citep{zhang2024contrastingintramodalrankingcrossmodal} because
 requires the VLM to be able to detect out-of-focus objects in low-resolution images \citep{why-winoground}.
In contrast, we propose to use FG-OVD \citep{FG-OVD}, a benchmark originally proposed to evaluate the ability of open-vocabulary object detectors to discern fine-grained object properties. 
In FG-OVD, negative captions are created starting from the object-specific captions by replacing attributes referring to the object's color, material, texture, etc.
We  crop the objects' bounding boxes which we use jointly with positive and negative captions and an image-to-text retrieval task  (more details in \Cref{app.fg_ovd}).



\begin{table*}[!ht]
    \caption{Comparison  between different generative training strategies. The value $\textbf{\textcolor{ForestGreen}{+x}}$ reported in the $i$-th row, column Average, refers to the average improvement across all datasets with respect to the method in row $i-1$.}
    \vspace{0.3em}
    \label{tab.Ablation-generative}
    \centering
    \setlength{\tabcolsep}{.15em}
    \resizebox{0.9\textwidth}{!}{%
    \begin{tabular}{lcccc c ccc c cccc c cccc c c c c c}
    \toprule
    & & \multicolumn{4}{c}{\textbf{ARO}} & \multicolumn{4}{c}{\textbf{SugarCrepe}} & & \multicolumn{4}{c}{\textbf{VL-Checklist}}  & & \textbf{ColorSwap} & &  \textbf{FG-OVD} & & \textbf{Avg}\\
    \cmidrule{3-5} \cmidrule{7-10} \cmidrule{12-15} \cmidrule{17-17} \cmidrule{19-19}  
     \textbf{Model} &  & \textbf{Relation} & \textbf{Attribute} & \textbf{Avg} & & \textbf{Add} & \textbf{Replace} & \textbf{Swap} & \textbf{Avg} & & \textbf{Attribute} & \textbf{Object} & \textbf{Relation} & \textbf{Avg} & & \textbf{ITT} & & \textbf{Avg} \\
     \midrule   
     Fully-Parallel & & 76.37 & 49.24 & 62.81 & & 98.98 & 77.98 & 68.81 & 81.92 & & 83.66 & 67.55 & 74.7 & 75.3 & & 25.24 & & 41.84 & & \textbf{57.42} \\
     Mixed & & 84.83 & 69.12 & 76.98 & & 99.01 & 84.17 & 78.39 & 87.19 & & 85.89 & 76.96 & 91.44 & 84.76 & & 41.33 & & 45.21 & & \hspace{17pt}\textbf{67.10}$_{\textbf{\textcolor{ForestGreen}{+9.67}}}$ \\        
     Sequential-AR & & 84.86 & 77.87 & 81.37 & & 98.96 & 83.62 & 81.50 & 88.02 & & 86.82 & 75.45 & 91.11 & 84.45 & & 46.33 & & 46.24 & & \hspace{17pt}\textbf{69.28}$_{\textbf{\textcolor{ForestGreen}{+2.18}}}$  \\
    \ours & & 87.56 & 90.26 & 88.91 & & 98.26 & 87.10 & 83.14 & 89.50 & & 86.07 & 78.91 & 89.37 & 84.78 & & 61.33 & & 51.48 & & \hspace{17pt}\textbf{75.20}$_{\textbf{\textcolor{ForestGreen}{+5.92}}}$  \\
    \bottomrule
\end{tabular}%
}
\end{table*}

\subsection{Ablations}
\label{sec.Ablation}

In the experiments of this section we 
follow a widely adopted protocol, first proposed in \citep{NegCLIP}, 
in which the VLM backbone is CLIP and the only training dataset is COCO \citep{coco}. However, we {\em do not} use the hard negatives of \citep{NegCLIP} for training because of their frequent semantic and syntactic errors (see above).
In \Cref{tab.Ablation-generative} we compare to each other the different word prediction strategies  described in \Cref{sec.Method}
using CLIP as the VLM.
 Specifically, 
 we indicate with {\em Sequential-AR} 
the replacement of  our decoder with  a standard AR decoder  (cross-attention over ${\cal Z}$ and standard causal attention over the past words of the caption), trained using the common
 image captioning objective defined in \Cref{eq.jointdistrib-AR}. Similarly to \ours, 
 we freeze the CLIP encoder and we use the visual features (${\cal Z}$) extracted from both the last and the penultimate layer of
 ${\cal E}$ (\Cref{sec.Decoder}). We use a same-size decoder, which takes only visible words as input (details in \Cref{app.ImplementationDetails}). {\em Sequential-AR} can be considered as our re-implementation of Cap\footnote{There are no publicly available network weights for \citep{Cap}.} \citep{Cap}, trained on COCO  and with a frozen visual encoder, which can directly be compared with the CGM-based strategy of \ours. 
 Similarly, we indicate with {\em Fully-Parallel} our re-implementation of the parallel prediction strategy proposed in \citep{Cap}, using a decoder which takes as input only masked tokens (only cross-attention over ${\cal Z}$  with a frozen visual encoder), trained using \Cref{eq.jointdistrib-parallel}.  Finally, in {\em Mixed} we use the sequential-parallel mixed strategy adopted in CapPa, in which, following \citep{Cap}, we use $75\%$ of the training samples with a parallel prediction (\Cref{eq.jointdistrib-parallel}) and $25\%$ of the samples with an AR prediction (\Cref{eq.jointdistrib-AR}). Architectural details are provided in \Cref{app.ImplementationDetails}.
 The results show that \ours outperforms all the other prediction strategies in all the datasets, often with a significant margin.
 For instance, \ours 
  achieves an average accuracy improvement of $+17.77$ points across all datasets
 with respect to {\em Fully-Parallel}, which arguably shows that the conditional independence assumption in \Cref{eq.jointdistrib-parallel} is too strong.
Overall,  
 these results confirm that
 an  off-the-shelf dependency parser provides  a priori knowledge 
 which can be exploited to model the conditional dependencies between  words in a sentence.

We further investigate the role of the dependency parser in \Cref{tab.Ablation-all-the-rest}.
Specifically,
 the column {\em Parser} refers to 
 the adopted dependency parser, where we compare 3 different methods: Deep Biaffine \citep{OurDependencyParser}, CRFPar \citep{zhang-etal-2020-efficient} and Deep Biaffine + RoBERTa \citep{OurDependencyParser}. 
 Note that we use the parsers as black boxes, without any training or fine-tuning, and the differences in the corresponding rows
of \Cref{tab.Ablation-all-the-rest} are based only in the use of a different external parser for \ours. 
 \Cref{tab.Ablation-all-the-rest} shows that the best results correspond to the use of Deep Biaffine + RoBERTa \citep{OurDependencyParser}, 
which is aligned with the higher accuracy of this parser compared to the other two according to the linguistic leaderboard
Penn Tree Bank \citep{parser-ranking}.
Note also that, according to this widely adopted  parser ranking \citep{parser-ranking},
there are higher performing parsers (e.g., \citep{best_parser}), however their code is not publicly available or it is not easy to use. Thus, we opted for Deep Biaffine + RoBERTa (used in all the other experiments of this paper). However, the results in \Cref{tab.Ablation-all-the-rest} show that, using a better parser, \ours can most likely achieve even better results.

The  {\em Mask-Specific} column in \Cref{tab.Ablation-all-the-rest} indicates the use of a dedicated masked token for each of the 45 syntactic categories of $V$ (\Cref{sec.Decoder}), which is compared with a generic BERT-like masked token \citep{devlin-etal-2019-bert}. In the latter case, we use the same masked token initial embedding vector for all the $n$ masked tokens fed to ${\cal D}$
(replicated $n$ times),
thus dropping any conditioning on $S_j$ in \Cref{eq.jointdistrib-disentangled}.
The results in \Cref{tab.Ablation-all-the-rest} show that this corresponds to a $-2.69$ point drop in accuracy averaged across all five datasets.

Finally, the  {\em Layers} column in \Cref{tab.Ablation-all-the-rest} indicates the number of  layers of CLIP we use to extract the visual features ${\cal Z}$: 
{\em Layers} = 1 means only the last layer ($L$); {\em Layers} = 2 means that we use also the penultimate layer (\Cref{sec.Decoder}). 
When the last layer only is used, the average accuracy drop is $-4.75$, showing the importance of using lower-level features in compositionality tasks where the VLM needs to consider small, non-foreground objects. 

In \Cref{subsec.additional_ablations} we provide the complete version of \Cref{tab.Ablation-all-the-rest} with all the possible combinations between the values of {\em Parser},  {\em Mask-Specific} and {\em Layers}, which confirm the results shown here.

\subsection{Main experiments}
\label{sec.SOTA-exp}

\begin{table*}[!t]
    \caption{Empirical contribution of different components of \ours.}
    \vspace{0.3em}
    \label{tab.Ablation-all-the-rest}
    \centering
    \setlength{\tabcolsep}{.15em}
    \resizebox{0.9\textwidth}{!}{%
    \begin{tabular}{lcccc c ccc c cccc c cccc c c c c c}
    \toprule
    & & & & \multicolumn{4}{c}{\textbf{ARO}} & \multicolumn{4}{c}{\textbf{SugarCrepe}} & & \multicolumn{4}{c}{\textbf{VL-Checklist}}  & & \textbf{ColorSwap} & & \textbf{FG-OVD} & & \textbf{Avg}\\
    \cmidrule{5-7} \cmidrule{9-12} \cmidrule{14-17} \cmidrule{19-19} \cmidrule{21-21} 
     \textbf{Parser} & \textbf{Mask-Specific} & \textbf{Layers} & & \textbf{Relation} & \textbf{Attribute} & \textbf{Avg} & & \textbf{Add} & \textbf{Replace} & \textbf{Swap} & \textbf{Avg} & & \textbf{Attribute} & \textbf{Object} & \textbf{Relation} & \textbf{Avg} & & \textbf{ITT} & & \textbf{Avg} \\
    \midrule
    CRFPar & \cmark & 2 & & 85.68 & 88.34 & 87.01 & & 98.16 & 84.94 & 80.30 & 87.80 & & 86.99 & 77.68 & 87.09 & 83.92 & & 56.33 & & 43.74 & & 71.76 \\
    Deep Biaffine & \cmark & 2 & & 86.56 & 89.10 & 87.83 & & 98.11 & 85.80 & 81.49 & 88.46 & & 87.02 & 78.30 & 87.75 & 84.35 & & 61.33 & & 44.74 & & 73.34  \\
    Deep Biaffine + RoBERTa & \xmark & 2 & & 84.75 & 86.16 & 85.46 & & 98.86 & 84.37 & 80.25 & 87.82 & & 83.79 & 78.24 & 90.84 & 84.29 & & 58.00 & & 46.99 & & 72.51 \\
    Deep Biaffine + RoBERTa & \cmark & 1 & & 86.82 & 89.67 & 88.25 & & 98.26 & 86.56 & 82.33 & 89.05 & & 84.41 & 78.94 & 89.54 & 84.30 & & 45.00 & & 45.63 & & 70.45 \\
    Deep Biaffine + RoBERTa & \cmark & 2 & & 87.56 & 90.26 & 88.91 & & 98.26 & 87.10 & 83.14 & 89.50 & & 86.07 & 78.91 & 89.37 & 84.78 & & 61.33 & & 51.48 & & 75.20  \\
    \bottomrule
\end{tabular}%
}
\end{table*}

{\bf Setting.} In this section we compare \ours with state-of-the-art compositional methods. 
Since different works are based on different VLMs and use different training data, to make the comparison as fair as possible, we split our evaluation based on both the VLM backbone and the used training set. Specifically,
in \Cref{tab.SOTA-comparison} we group the approaches based on CLIP \citep{radford2021learning} and in  \Cref{tab.SOTA-comparison-xvlm} those  which adopt a different VLM,
while  \Cref{tab.SOTA-comparison-instructblip} is dedicated to methods based on VLMs pre-trained using a language based decoder. In the first category, our approach is indicated by  \ours-CLIP.
In the second group, we use XVLM \citep{xvlm} as our backbone (\ours-XVLM). Finally, we use InstructBLIP \citep{instructblip} for the VLM category with a language-based decoder  (\ours-InstructBLIP) (more details in \Cref{app.ImplementationDetails}). 
\ours-CLIP, \ours-XVLM and \ours-InstructBLIP 
are trained on COCO {\em  only} 
($\sim 100$K training samples, see \Cref{sec.Ablation}).
Moreover, following \citep{zhang2024contrastingintramodalrankingcrossmodal}, we present additional results training on a combination of three datasets:
 COCO, CC3M \citep{cc3m}, and Visual Genome \citep{visual_genome}, and we call the corresponding methods as 
\ours-CLIP+,  \ours-XVLM+ and \ours-InstructBLIP+.
In this case, we use a decoder ${\cal D}$ with four blocks.
Note that we use only $\sim50K$ samples from Visual Genome because we {\em removed those training data which overlap with ARO and VL-Checklist}. 
On the other hand, CC3M 
($\sim$$3.3$M training samples) is a much larger but also noisier dataset, since its
captions are obtained from the Alt-text HTML attribute associated with web images, and we use it also to indirectly evaluate the robustness of \ours to noisy textual descriptions (see \Cref{subsec.additional_ablations}).
For each compared baseline, the results shown in the tables refer to the values reported in the original article (when available) or to our reproduction using the (possibly available) public code.
The results on FG-OVD are averaged over all tasks and we report in \Cref{subsec.fg-ovd} the task-specific  values.

\begin{table*}[!ht]
    \caption{Comparison with  compositional methods based on CLIP.
    For each baseline, we report the values published in the original article. In case a given dataset was not used by that baseline, but a public code is available, we report the results obtained by our reproduction.
    With     $\textbf{x}$, $\underline{x}$ and $x^*$ 
    we indicate the first, the second and the third best result, respectively.}
    \vspace{0.2em}
    \label{tab.SOTA-comparison}
    \centering
    \setlength{\tabcolsep}{.15em}
    \resizebox{0.9\textwidth}{!}{%
    \begin{tabular}{lcccc c ccc c cccc c cccc c c c c c}
    \toprule
    & & \multicolumn{3}{c}{\textbf{ARO}} & & \multicolumn{4}{c}{\textbf{SugarCrepe}} & & \multicolumn{4}{c}{\textbf{VL-Checklist}}  & & \textbf{ColorSwap} & & \textbf{FG-OVD} & & \textbf{Avg} \\
    \cmidrule{3-5} \cmidrule{7-10} \cmidrule{12-15} \cmidrule{17-17} \cmidrule{19-19}
     \textbf{Model} & & \textbf{Relation} & \textbf{Attribute} & \textbf{Avg} & & \textbf{Add} & \textbf{Replace} & \textbf{Swap} & \textbf{Avg} & & \textbf{Attribute} & \textbf{Object} & \textbf{Relation} & \textbf{Avg} & & \textbf{ITT} & &  \textbf{Avg} \\
    \midrule
    \multicolumn{1}{c}{\textit{Zero-shot}} \\
    CLIP~\citep{radford2021learning} & & 59.00 & 62.00 & 60.50 & & 85.58 & 80.76  & 70.83 & 79.05 & & 67.93 & 82.83 & 64.19 & 71.65 & & 35.67\rlap{\raisebox{0.5ex}{*}} & & 47.33 & & 58.84 \\
    \midrule
    \multicolumn{1}{c}{\textit{Training on COCO only}} \\
    CLIP Fine-Tuned~\citep{NegCLIP} & & 63.00 & 65.00 & 64.00 & & . & .  & . & .  & & . & . & . & . & & . & & . & & . \\
    NegCLIP~\citep{NegCLIP} & & 81.00 & 71.00 & 76.00 & & 87.29 & 85.36  & 75.30 & 82.65  & & 72.24 & 87.00 & 71.39 & 76.87 & & 35.67\rlap{\raisebox{0.5ex}{*}} & & 41.69 & & 62.57 \\
    CE-CLIP~\citep{zhang2024contrastingintramodalrankingcrossmodal} & & 83.00 & 76.40 & 79.70 & & 92.90 & 87.00 & 74.90 & 84.94 & & 72.60 & 84.60 & 71.80 & 76.30 & & 18.67 & & 41.97 & & 60.31 \\
    Structure-CLIP~\citep{huang2023structureclipscenegraphknowledge} & & 85.10\rlap{\raisebox{0.5ex}{*}} & 83.50\rlap{\raisebox{0.5ex}{*}} & 84.30\rlap{\raisebox{0.5ex}{*}} & & . & . & . & . & & . & . & . & . & & . & & . & & . \\ GNM~\citep{sahin2023enhancingmultimodalcompositionalreasoning} & & 65.00 & 65.00 & 65.00 & & 82.85 & 80.95 & 66.71 & 76.83 & & 70.15 & 85.91 & 64.10 & 73.38 & & 13.00 & & 38.79 & & 53.40 \\
    Plausible Adj. Neg~\citep{buettner2023investigatingroleattributecontext} & & 65.07 & 67.94 & 66.51 & & 89.64 & 85.37 & 70.88 & 81.96 & & 76.51 & \underline{88.13} & 69.90 & 78.17 & & 17.67 & & 44.98 & & 57.86 \\
    SDS-CLIP~\citep{SDS-CLIP} & & 55.00 & 66.00 & 60.50 & & . & . & . & . & & . & . & . & . & & . & & . & & . \\
    \rowcolor[gray]{0.9}
    \ours-CLIP & & \underline{87.56} & \underline{90.26} & \underline{88.91} & & \underline{98.26} & 87.10\rlap{\raisebox{0.5ex}{*}} & \underline{83.14} & \underline{89.50} & & \underline{86.07} & 78.91 & 89.37\rlap{\raisebox{0.5ex}{*}} & \underline{84.78} & & \underline{61.33} & & \underline{51.48} & & \underline{75.20} \\
    \midrule
    \multicolumn{1}{c}{\textit{Training on datasets larger than COCO}} \\
    CE-CLIP+~\citep{zhang2024contrastingintramodalrankingcrossmodal} & & 83.60 & 77.10 & 80.35 & & 94.40 & \textbf{89.30} & 78.00\rlap{\raisebox{0.5ex}{*}} & 87.23\rlap{\raisebox{0.5ex}{*}} & & 76.70 & 86.30 & 74.70 & 79.23 & & . & & . & & . \\
    IL-CLIP~\citep{zheng2024iteratedlearningimprovescompositionality} & & . & . & . & & 73.80 & 73.00 & 62.90 & 69.90 & & . & . & . & . & & . & & . & & . \\
    syn-CyCLIP~\citep{cascantebonilla2023goingnounsvision} & & 69.00 & 63.65 & 66.33 & & . & . & . & . & & 68.06 & . & 65.73 & . & & . & & . & & . \\
    CLIP-SPEC~\citep{clip-spec} & & 73.70 & 66.40 & 70.05 & & . & . & . & . & & . & . & . & . & & . & & . & & \\
    DAC-SAM~\citep{DAC} & & 77.16 & 70.50 & 73.83 & & 92.87 & 86.18 & 71.06 & 83.37 & & 75.80 & \textbf{88.50} & \underline{89.80} & 84.70\rlap{\raisebox{0.5ex}{*}} & & 16.33 & & 48.36 & & 61.31 \\
    DAC-LLM~\citep{DAC} & & 81.28 & 73.91 & 77.60 & & 95.83\rlap{\raisebox{0.5ex}{*}} & \underline{88.09} & 72.48 & 85.47 & & 77.30\rlap{\raisebox{0.5ex}{*}} & 87.30\rlap{\raisebox{0.5ex}{*}} & 86.40 & 83.66 & & 18.33 & & 49.60\rlap{\raisebox{0.5ex}{*}} & & 62.93\rlap{\raisebox{0.5ex}{*}} \\
    \rowcolor[gray]{0.9}
    \ours-CLIP+ & & \textbf{90.67} & \textbf{96.01} & \textbf{93.34} & & \textbf{98.42} & 87.05 & \textbf{84.21} & \textbf{89.89} & & \textbf{90.71} & 84.91 & \textbf{92.33} & \textbf{89.31} & & \textbf{81.66} & & \textbf{69.96} & & \textbf{84.83} \\    
    \bottomrule
\end{tabular}%
}
\end{table*}

{\bf CLIP based methods.}
\Cref{tab.SOTA-comparison} shows that \ours-CLIP 
{\em largely} outperforms all the other  approaches trained only on COCO 
{\bf and it also outperforms all the  methods trained on datasets larger than COCO}. 
For instance, using the average across all the datasets, 
\ours-CLIP outperforms the second best result in \Cref{tab.SOTA-comparison} (DAC-LLM \citep{DAC}) by a remarkable $12.27$ points. Note that DAC-LLM was trained on CC3M, 
a  dataset an order of magnitude larger than COCO,
and using high-quality LLM-based annotations  (\Cref{sec.RelatedWork}). 
Moreover, even considering the average computed on the individual datasets, \ours-CLIP
outperforms all the other methods in \Cref{tab.SOTA-comparison} (including  those trained on datasets larger than COCO).
We believe that these results show that  our CGM-based training strategy 
can better generalize  by leveraging available training data, most likely because we  remove spurious inter-variable associations from the learning objective (\Cref{sec.Method}).
Moreover, \ours-CLIP+  
achieves even better results, with an average across all benchmarks that is almost 22 points more than the second best result (DAC-LLM).

\begin{table*}[!ht]
    \caption{Comparison with  methods based on other VLMs.     
    Similar to \Cref{tab.SOTA-comparison}, the baseline results are either taken from the original paper or reproduced using the public code.}
    \vspace{0.2em}
    \label{tab.SOTA-comparison-xvlm}
    \centering
    \setlength{\tabcolsep}{.15em}
    \resizebox{0.9\textwidth}{!}{%
    \begin{tabular}{lcccc c ccc c cccc c cccc c c c c c}
    \toprule
    & & \multicolumn{3}{c}{\textbf{ARO}} & & \multicolumn{4}{c}{\textbf{SugarCrepe}} & & \multicolumn{4}{c}{\textbf{VL-Checklist}}  & & \textbf{ColorSwap} & & \textbf{FG-OVD} & &  \textbf{Avg} \\
    \cmidrule{3-5} \cmidrule{7-10} \cmidrule{12-15} \cmidrule{17-17} \cmidrule{19-19}
     \textbf{Model} & & \textbf{Relation} & \textbf{Attribute} & \textbf{Avg} & & \textbf{Add} & \textbf{Replace} & \textbf{Swap} & \textbf{Avg} & & \textbf{Attribute} & \textbf{Object} & \textbf{Relation} & \textbf{Avg} & & \textbf{ITT} & &  \textbf{Avg} \\
    \midrule
    \multicolumn{1}{c}{\textit{Zero-shot}} \\
    XVLM~\citep{xvlm} & & 73.40 & 86.80 & 80.10 & & . & . & . & . & & 75.10\rlap{\raisebox{0.5ex}{*}} & \underline{85.80} & 70.40 & 76.50 & & . & & . & & . \\
    \midrule
    \multicolumn{1}{c}{\textit{Training on COCO only}} \\
    CE-XVLM~\citep{zhang2024contrastingintramodalrankingcrossmodal} & & 73.90\rlap{\raisebox{0.5ex}{*}} & 89.30\rlap{\raisebox{0.5ex}{*}} & 81.60\rlap{\raisebox{0.5ex}{*}} & & . & . & . & . & & 74.80 & \textbf{86.90} & 79.70\rlap{\raisebox{0.5ex}{*}} & 78.60\rlap{\raisebox{0.5ex}{*}} & & . & & . & & . \\
    HardNeg-DiffusionITM~\citep{krojer2023diffusionmodelsvisionandlanguagereasoners} & & 52.30 & 67.60 & 59.95 & & . & . & . & . & & . & . & . & . & & . & & . & & . \\
    \rowcolor[gray]{0.9}
    \ours-XVLM & & \underline{87.64} & \underline{92.30} & \underline{89.97} & & \textbf{98.65}\rlap{\raisebox{0.5ex}{*}} & \textbf{89.17} & \underline{84.37} & \underline{90.73} & & \underline{85.87} & 80.49 & \underline{88.74} & \underline{85.03} & & \underline{69.67} & & \underline{50.12} & & \underline{77.10}\\
    \midrule
    \multicolumn{1}{c}{\textit{Training on datasets larger than COCO}} \\
    \rowcolor[gray]{0.9}
    \ours-XVLM+ & & \textbf{91.71} & \textbf{96.59} & \textbf{94.15} & & \underline{98.30} & \underline{88.97} & \textbf{86.49} & \textbf{91.25} & & \textbf{91.54} & 84.73\rlap{\raisebox{0.5ex}{*}} & \textbf{92.33} & \textbf{89.53} & & \textbf{82.33} & & \textbf{74.22} & & \textbf{86.30}\\

    \bottomrule
\end{tabular}%
}
\end{table*}

\begin{table*}[!ht]
    \caption{Comparison with  methods based on encoder-decoder VLM architectures pre-trained with a textual token prediction task. \textsuperscript{\dag}~\citet{Revisiting} show additional results with $\alpha$ set using dataset-specific cross-validation data, which we do not report, however, to make the comparison fair to other methods that do not have access to benchmark data.}
    \vspace{0.2em}
    \label{tab.SOTA-comparison-instructblip}
    \centering
    \setlength{\tabcolsep}{.15em}
    \resizebox{0.9\textwidth}{!}{%
    \begin{tabular}{lcccc c ccc c cccc c cccc c c c c c}
    \toprule
    & & \multicolumn{3}{c}{\textbf{ARO}} & & \multicolumn{4}{c}{\textbf{SugarCrepe}} & & \multicolumn{4}{c}{\textbf{VL-Checklist}}  & & \textbf{ColorSwap} & & \textbf{FG-OVD} & &  \textbf{Avg} \\
    \cmidrule{3-5} \cmidrule{7-10} \cmidrule{12-15} \cmidrule{17-17} \cmidrule{19-19}
     \textbf{Model} & & \textbf{Relation} & \textbf{Attribute} & \textbf{Avg} & & \textbf{Add} & \textbf{Replace} & \textbf{Swap} & \textbf{Avg} & & \textbf{Attribute} & \textbf{Object} & \textbf{Relation} & \textbf{Avg} & & \textbf{ITT} & &  \textbf{Avg} \\
    \midrule
    BLIP~\citep{blip} & & 59.00 & 88.00 & 73.50 & & . & . & . & . & & 75.20 & 82.20 & 70.50 & 75.70 & & . & & . & & . \\
    BLIP2~\citep{blip2} & & 41.20 & 71.30 & 56.25 & & . & . & . & . & & 77.80 & 84.90 & 70.60 & 77.80 & & . & & . & & . \\
    InstructBLIP (FlanT5XL)~\citep{instructblip} & & 69.20 & 50.83 & 60.02 & & 65.43 & 72.59 & 63.41 & 67.14 & & 56.37 & 80.33 & 53.34 & 63.35 & & 40.33\rlap{\raisebox{0.5ex}{*}} & & 26.80\rlap{\raisebox{0.5ex}{*}} & & 51.53\rlap{\raisebox{0.5ex}{*}} \\
    MiniGPT-4~\citep{minigpt4} & & 46.90 & 55.70 & 51.30 & & . & . & . & . & & 71.30 & 84.20 & . & . & & . & & . & & . \\
    GPT-4V~\citep{gpt4v} & & . & . & . & & 91.68 & \textbf{93.37} & 86.61 & 90.55 & & . & . & . & . & & . & & . & & . \\
    LLaVA-1.5-13B~\citep{LLaVA} & & . & . & . & & . & . & 80.95 & . & & . & . & . & . & & . & & . & & . \\
    LLaVA-1.5-13B+CRG~\citep{CRG} & & . & . & . & & . & . & 87.90 & . & & . & . & . & . & & . & & . & & . \\
    LLaVA-1.6-34B~\citep{llavanext} & & . & . & . & & . & . & 81.25 & . & & . & . & . & . & & . & & . & & . \\
    LLaVA-1.6-34B+CRG~\citep{CRG} & & . & . & . & & . & . & \underline{90.75} & . & & . & . & . & . & & . & & . & & . \\
    BLIP-VisualGPTScore ($\alpha = 0$)~\citep{Revisiting} \dag & & 89.10\rlap{\raisebox{0.5ex}{*}} & \underline{95.30} & 92.20\rlap{\raisebox{0.5ex}{*}} & & 91.00 & \underline{93.30} & \textbf{91.00} & 91.77\rlap{\raisebox{0.5ex}{*}} & & 78.70\rlap{\raisebox{0.5ex}{*}} & \textbf{92.60} & \underline{90.80} & \underline{87.37} & & . & & . & & . \\
    BLIP2-VisualGPTScore ($\alpha = 0$)~\citep{Revisiting} \dag & & \underline{90.70} & 94.30\rlap{\raisebox{0.5ex}{*}} & \underline{92.50} & & 92.70 & 93.00\rlap{\raisebox{0.5ex}{*}} & 91.24 & \underline{92.31} & & 73.90 & \underline{90.10} & 89.90\rlap{\raisebox{0.5ex}{*}} & 84.63 & & . & & . & & . \\
    Cap~\citep{Cap}  & & 86.60 & 88.90 & 87.75 & & \underline{98.94} & 88.21 & 84.00 & 90.38 & & . & . & . & . & & . & & . \\
    CapPa~\citep{Cap} & & 86.70 & 85.70 & 86.20 & & \textbf{99.13} & 87.67 & 83.11 & 89.97 & & . & . & . & . & & . & & . & & . \\
    \midrule
    \rowcolor[gray]{0.9}
    \ours-InstructBLIP & & 87.63 & 88.93 & 88.28 & & 98.55\rlap{\raisebox{0.5ex}{*}} & 90.61 & 88.12 & \textbf{92.42} & & \underline{85.77} & 79.96 & 89.14 & 84.96\rlap{\raisebox{0.5ex}{*}} & & \underline{72.66} & & \underline{51.26} & & \underline{77.87} \\
    \rowcolor[gray]{0.9}
    \ours-InstructBLIP+ & & \textbf{91.12} & \textbf{95.64} & \textbf{93.38} & & 98.45 & 90.27 & 88.22\rlap{\raisebox{0.5ex}{*}} & \underline{92.31} & & \textbf{90.80} & 85.17\rlap{\raisebox{0.5ex}{*}} & \textbf{92.80} & \textbf{89.60} & & \textbf{83.33} & & \textbf{70.72} & & \textbf{85.87} \\
    \bottomrule
\end{tabular}%
}
\end{table*}

\begin{table*}[!ht]
    \caption{Comparison of CLIP-based models using image classification tasks and linear probing.}
    \vspace{0.2em}
    \label{tab:linear-probing}
    \centering
    \setlength{\tabcolsep}{.15em}
    \resizebox{0.6\columnwidth}{!}{%
    \begin{tabular}{lcccc}
    \toprule
    \textbf{Model} & \textbf{CIFAR10} & \textbf{CIFAR100} & \textbf{ImageNet1K (top 1)} & \textbf{ImageNet1K (top 5)} \\
    \midrule
    CLIP~\citep{radford2021learning} & 94.2 & 79.0 & \underline{75.0} & 93.2 \\
    CLIP Fine-Tuned~\citep{NegCLIP} & 95.0 & 80.0 & 74.0 & - \\
    NegCLIP~\citep{NegCLIP} & 94.0 & 79.0 & 72.0 & - \\
    CE-CLIP~\citep{zhang2024contrastingintramodalrankingcrossmodal} & 93.8 & 78.0 & - & 92.6 \\
    CE-CLIP+~\citep{zhang2024contrastingintramodalrankingcrossmodal} & 93.8 & 78.1 & - & 92.7 \\
    \rowcolor[gray]{0.9}
    COGT-CLIP & \underline{96.7} & \underline{84.9} & 74.4 & \underline{93.3} \\
    \rowcolor[gray]{0.9}
    COGT-CLIP+ & \textbf{96.8} & \textbf{85.4} & \textbf{75.3} & \textbf{93.8} \\
    \bottomrule
    \end{tabular}%
    }
\end{table*}

{\bf Other VLMs.}
The results 
in \Cref{tab.SOTA-comparison}  are confirmed by those reported in \Cref{tab.SOTA-comparison-xvlm},
where 
\ours-XVLM largely outperforms the other methods, included CE-XVLM \citep{zhang2024contrastingintramodalrankingcrossmodal}, which uses our same VLM (XVLM) and the same training data (COCO).
 \ours-XVLM+ 
further improves these results and it also outperforms \ours-CLIP+. This is probably because the XVLM encoder
can better represent small-scale objects than the CLIP encoder, and these objects are often referenced in the captions of
these compositional benchmarks \citep{NegCLIP}. 

{\bf Language-decoding based  VLMs.} 
In \Cref{tab.SOTA-comparison-instructblip} we compare to each other VLMs  pre-trained using a decoder and a generative word prediction task. The compositional skills of these methods are generally much higher than the other VLMs,
which indirectly confirms that a word-prediction training helps the VLM to understand the compositional nature of the human language (\Cref{sec.RelatedWork}). However, 
a direct comparison with VLMs such as CLIP, XVLM or Stable Diffusion \citep{krojer2023diffusionmodelsvisionandlanguagereasoners} is difficult, since each of these backbones has been pre-trained on datasets with a  huge difference in size.
For instance,  Cap and CapPa \citep{Cap} 
were pre-trained with a private dataset composed of 1B image/Alt-text pairs \citep{Cap}, which is different orders of magnitude larger  than  the dataset 
used to pre-train XVLM  ($\sim 16$M  training samples \citep{xvlm}).
 Despite that, \ours-XVLM+ (\Cref{tab.SOTA-comparison-xvlm}) outperforms both Cap and CapPa and all the other methods in \Cref{tab.SOTA-comparison-instructblip}. Similarly,
 \ours-InstructBLIP+ significantly outperforms the zero-shot accuracy of InstructBLIP and, jointly with \ours-XVLM+,
establishes {\em a new state of the art on each of these compositional datasets}.

 \subsection{Downstream tasks}

\citet{Doveh_2023_CVPR} show that most
compositional methods usually deteriorate the VLM skills on non-compositional, standard tasks. We analyze this aspect  using the protocol adopted by \citep{NegCLIP,zhang2024contrastingintramodalrankingcrossmodal}, which is based on 
linear probing  the fine-tuned CLIP visual encoder on CIFAR10, CIFAR100 and ImageNet. Since in \ours ${\cal E}$ is frozen,  we use ${\cal E}$ jointly with our mapping network ${\cal M}$ and, specifically, the
feature $\pmb{z}_{[\texttt{CLS}]}^L$. The results shown in \Cref{tab:linear-probing} show that COGT 
not only does it not deteriorate CLIP's features but it can even improve them.

\section{Conclusion}
\label{sec.Conclusion}

In this paper we presented \ours, a compositional method based on a semi-parallel generative training. Specifically, we exploit the a priori knowledge of an off-the-shelf  dependency parser to define a set of  causal relations between the words of a sentence. These relations,
collectively represented using a CGM, are used to 
make sparser  the joint probability distribution of the textual variables by removing possible spurious inter-variable associations. As a result, \ours can better exploit the training data and reduce the risk of overfitting.
Using extensive experiments, we showed that \ours is much more effective than standard AR or fully-parallel generative predictions and it largely outperforms all previous compositional works, including methods trained with much larger datasets.

\subsubsection*{Acknowledgments}

This work was supported by the PNRR project ``Italian Strengthening of Esfri RI Resilience (ITSERR)'', funded by the European Union – NextGenerationEU (CUP B53C22001770006),
 by the EU Horizon project ELIAS (No.
101120237), and by the PNRR project  ``A Static and Dynamic Database for Historic Urban Contexts Accessibility'', funded by the European Union – NextGenerationEU (CUP E53D23010500001).

\bibliography{iclr2025_conference}

\begin{thebibliography}{92}
\providecommand{\natexlab}[1]{#1}
\providecommand{\url}[1]{\texttt{#1}}
\expandafter\ifx\csname urlstyle\endcsname\relax
  \providecommand{\doi}[1]{doi: #1}\else
  \providecommand{\doi}{doi: \begingroup \urlstyle{rm}\Url}\fi

\bibitem[Basu et~al.(2023)Basu, Sanjabi, Massiceti, Hu, and Feizi]{SDS-CLIP}
Samyadeep Basu, Maziar Sanjabi, Daniela Massiceti, Shell~Xu Hu, and Soheil Feizi.
\newblock Augmenting {CLIP} with improved visio-linguistic reasoning.
\newblock In \emph{arXiv:2307.09233}, 2023.

\bibitem[Basu et~al.(2024)Basu, Hu, Sanjabi, Massiceti, and Feizi]{Distilling}
Samyadeep Basu, Shell~Xu Hu, Maziar Sanjabi, Daniela Massiceti, and Soheil Feizi.
\newblock Distilling knowledge from text-to-image generative models improves visio-linguistic reasoning in {CLIP}.
\newblock In \emph{arXiv:2307.09233}, 2024.

\bibitem[Bianchi et~al.(2024)Bianchi, Carrara, Messina, Gennaro, and Falchi]{FG-OVD}
Lorenzo Bianchi, Fabio Carrara, Nicola Messina, Claudio Gennaro, and Fabrizio Falchi.
\newblock The devil is in the fine-grained details: Evaluating open-vocabulary object detectors for fine-grained understanding.
\newblock In \emph{CVPR}, 2024.

\bibitem[Bolelli et~al.(2018)Bolelli, Baraldi, Pollastri, and Grana]{bolelli2018hierarchical}
Federico Bolelli, Lorenzo Baraldi, Federico Pollastri, and Costantino Grana.
\newblock A hierarchical quasi-recurrent approach to video captioning.
\newblock In \emph{2018 IEEE International Conference on Image Processing, Applications and Systems (IPAS)}, 2018.

\bibitem[Buettner \& Kovashka(2024)Buettner and Kovashka]{buettner2023investigatingroleattributecontext}
Kyle Buettner and Adriana Kovashka.
\newblock Investigating the role of attribute context in vision-language models for object recognition and detection.
\newblock In \emph{WACV}, 2024.

\bibitem[Burapacheep et~al.(2024)Burapacheep, Gaur, Bhatia, and Thrush]{colorswap}
Jirayu Burapacheep, Ishan Gaur, Agam Bhatia, and Tristan Thrush.
\newblock Colorswap: A color and word order dataset for multimodal evaluation.
\newblock In \emph{arXiv:2402.04492}, 2024.

\bibitem[Cascante-Bonilla et~al.(2023)Cascante-Bonilla, Shehada, Smith, Doveh, Kim, Panda, Varol, Oliva, Ordonez, Feris, and Karlinsky]{cascantebonilla2023goingnounsvision}
Paola Cascante-Bonilla, Khaled Shehada, James~Seale Smith, Sivan Doveh, Donghyun Kim, Rameswar Panda, Gül Varol, Aude Oliva, Vicente Ordonez, Rogerio Feris, and Leonid Karlinsky.
\newblock Going beyond nouns with vision \& language models using synthetic data.
\newblock In \emph{ICCV}, 2023.

\bibitem[Chen \& Manning(2014)Chen and Manning]{chen-manning-2014-fast}
Danqi Chen and Christopher Manning.
\newblock A fast and accurate dependency parser using neural networks.
\newblock In \emph{Conference on Empirical Methods in Natural Language Processing ({EMNLP})}, 2014.

\bibitem[Clark \& Jaini(2023)Clark and Jaini]{clark2023texttoimagediffusionmodelszeroshot}
Kevin Clark and Priyank Jaini.
\newblock Text-to-image diffusion models are zero-shot classifiers.
\newblock In \emph{arXiv:2303.15233}, 2023.

\bibitem[Dai et~al.(2023)Dai, Li, Li, Tiong, Zhao, Wang, Li, Fung, and Hoi]{instructblip}
Wenliang Dai, Junnan Li, Dongxu Li, Anthony Tiong, Junqi Zhao, Weisheng Wang, Boyang Li, Pascale Fung, and Steven Hoi.
\newblock Instruct{BLIP}: Towards general-purpose vision-language models with instruction tuning.
\newblock In \emph{NeurIPS}, 2023.

\bibitem[Deguchi et~al.(2019)Deguchi, Tamura, and Ninomiya]{Deguchi2019DependencyBasedSF}
Hiroyuki Deguchi, Akihiro Tamura, and Takashi Ninomiya.
\newblock Dependency-based self-attention for transformer {NMT}.
\newblock In \emph{Recent Advances in Natural Language Processing (RANLP)}, 2019.

\bibitem[Diaconescu(2002)]{Diaconescu}
Stefan Diaconescu.
\newblock A generative dependency grammar.
\newblock In \emph{PRICAI 2002: Trends in Artificial Intelligence}, 2002.

\bibitem[Diwan et~al.(2022)Diwan, Berry, Choi, Harwath, and Mahowald]{why-winoground}
Anuj Diwan, Layne Berry, Eunsol Choi, David Harwath, and Kyle Mahowald.
\newblock Why is winoground hard? investigating failures in visuolinguistic compositionality.
\newblock In \emph{Conference on Empirical Methods in Natural Language Processing (EMNLP)}, 2022.

\bibitem[Dosovitskiy et~al.(2021)Dosovitskiy, Beyer, Kolesnikov, Weissenborn, Zhai, Unterthiner, Dehghani, Minderer, Heigold, Gelly, Uszkoreit, and Houlsby]{ViT}
Alexey Dosovitskiy, Lucas Beyer, Alexander Kolesnikov, Dirk Weissenborn, Xiaohua Zhai, Thomas Unterthiner, Mostafa Dehghani, Matthias Minderer, Georg Heigold, Sylvain Gelly, Jakob Uszkoreit, and Neil Houlsby.
\newblock An image is worth 16x16 words: Transformers for image recognition at scale.
\newblock In \emph{ICLR}, 2021.

\bibitem[Doveh et~al.(2023{\natexlab{a}})Doveh, Arbelle, Harary, Herzig, Kim, Cascante-Bonilla, Alfassy, Panda, Giryes, Feris, Ullman, and Karlinsky]{DAC}
Sivan Doveh, Assaf Arbelle, Sivan Harary, Roei Herzig, Donghyun Kim, Paola Cascante-Bonilla, Amit Alfassy, Rameswar Panda, Raja Giryes, Rogerio Feris, Shimon Ullman, and Leonid Karlinsky.
\newblock Dense and aligned captions ({DAC}) promote compositional reasoning in {VL} models.
\newblock In \emph{NeurIPS}, 2023{\natexlab{a}}.

\bibitem[Doveh et~al.(2023{\natexlab{b}})Doveh, Arbelle, Harary, Schwartz, Herzig, Giryes, Feris, Panda, Ullman, and Karlinsky]{Doveh_2023_CVPR}
Sivan Doveh, Assaf Arbelle, Sivan Harary, Eli Schwartz, Roei Herzig, Raja Giryes, Rogerio Feris, Rameswar Panda, Shimon Ullman, and Leonid Karlinsky.
\newblock Teaching structured vision \& language concepts to vision \& language models.
\newblock In \emph{CVPR}, 2023{\natexlab{b}}.

\bibitem[Dozat \& Manning(2016)Dozat and Manning]{OurDependencyParser}
Timothy Dozat and Christopher~D. Manning.
\newblock Deep biaffine attention for neural dependency parsing.
\newblock In \emph{arXiv:1611.01734}, 2016.

\bibitem[Du et~al.(2020)Du, Lin, Shen, O'Donnell, Bengio, and Zhang]{du2020exploitingsyntacticstructurebetter}
Wenyu Du, Zhouhan Lin, Yikang Shen, Timothy~J. O'Donnell, Yoshua Bengio, and Yue Zhang.
\newblock Exploiting syntactic structure for better language modeling: A syntactic distance approach.
\newblock In \emph{ACL}, 2020.

\bibitem[Ghiasi et~al.(2022)Ghiasi, Kazemi, Borgnia, Reich, Shu, Goldblum, Wilson, and Goldstein]{visual_features_penultimate}
Amin Ghiasi, Hamid Kazemi, Eitan Borgnia, Steven Reich, Manli Shu, Micah Goldblum, Andrew~Gordon Wilson, and Tom Goldstein.
\newblock {What do vision transformers learn? A visual exploration}.
\newblock In \emph{arXiv:2212.06727}, 2022.

\bibitem[Goyal \& Bengio(2020)Goyal and Bengio]{Goyal2020InductiveBF}
Anirudh Goyal and Yoshua Bengio.
\newblock Inductive biases for deep learning of higher-level cognition.
\newblock \emph{Proceedings of the Royal Society A}, 478, 2020.

\bibitem[Goyal et~al.(2021)Goyal, Lamb, Hoffmann, Sodhani, Levine, Bengio, and Sch{\"o}lkopf]{goyal2021recurrent}
Anirudh Goyal, Alex Lamb, Jordan Hoffmann, Shagun Sodhani, Sergey Levine, Yoshua Bengio, and Bernhard Sch{\"o}lkopf.
\newblock Recurrent independent mechanisms.
\newblock In \emph{ICLR}, 2021.

\bibitem[Herzig et~al.(2023)Herzig, Mendelson, Karlinsky, Arbelle, Feris, Darrell, and Globerson]{Incorporating}
Roei Herzig, Alon Mendelson, Leonid Karlinsky, Assaf Arbelle, Rogerio Feris, Trevor Darrell, and Amir Globerson.
\newblock Incorporating structured representations into pretrained vision \& language models using scene graphs.
\newblock In \emph{Conference on Empirical Methods in Natural Language Processing (EMNLP)}, 2023.

\bibitem[Ho et~al.(2020)Ho, Jain, and Abbeel]{DDPMs}
Jonathan Ho, Ajay Jain, and Pieter Abbeel.
\newblock Denoising diffusion probabilistic models.
\newblock In \emph{NeurIPS}, 2020.

\bibitem[Honnibal et~al.(2020)Honnibal, Montani, Van~Landeghem, and Boyd]{honnibal2020spacy}
Matthew Honnibal, Ines Montani, Sofie Van~Landeghem, and Adriane Boyd.
\newblock {spaCy: Industrial-strength Natural Language Processing in Python}, 2020.

\bibitem[Hou et~al.(2022)Hou, Kai, Xue, Zhu, Yuan, Huang, Wang, and Lin]{hou2022syntaxguidedlocalizedselfattentionconstituency}
Shengyuan Hou, Jushi Kai, Haotian Xue, Bingyu Zhu, Bo~Yuan, Longtao Huang, Xinbing Wang, and Zhouhan Lin.
\newblock Syntax-guided localized self-attention by constituency syntactic distance.
\newblock In \emph{arXiv:2210.11759}, 2022.

\bibitem[Hsieh et~al.(2024)Hsieh, Zhang, Ma, Kembhavi, and Krishna]{sugarcrepe}
Cheng-Yu Hsieh, Jieyu Zhang, Zixian Ma, Aniruddha Kembhavi, and Ranjay Krishna.
\newblock {SugarCrepe:} fixing hackable benchmarks for vision-language compositionality.
\newblock In \emph{NeurIPS}, 2024.

\bibitem[Huang et~al.(2024)Huang, Tang, Chen, Zhang, Zhang, Chen, Zhao, Zhao, Lv, Hu, and Zhang]{huang2023structureclipscenegraphknowledge}
Yufeng Huang, Jiji Tang, Zhuo Chen, Rongsheng Zhang, Xinfeng Zhang, Weijie Chen, Zeng Zhao, Zhou Zhao, Tangjie Lv, Zhipeng Hu, and Wen Zhang.
\newblock {Structure-CLIP: T}owards scene graph knowledge to enhance multi-modal structured representations.
\newblock In \emph{AAAI}, 2024.

\bibitem[Ilharco et~al.(2021)Ilharco, Wortsman, Wightman, Gordon, Carlini, Taori, Dave, Shankar, Namkoong, Miller, Hajishirzi, Farhadi, and Schmidt]{open_clip}
Gabriel Ilharco, Mitchell Wortsman, Ross Wightman, Cade Gordon, Nicholas Carlini, Rohan Taori, Achal Dave, Vaishaal Shankar, Hongseok Namkoong, John Miller, Hannaneh Hajishirzi, Ali Farhadi, and Ludwig Schmidt.
\newblock Openclip, 2021.
\newblock URL \url{https://doi.org/10.5281/zenodo.5143773}.

\bibitem[Jiang et~al.(2024)Jiang, He, Xu, and Wang]{ComCLIP}
Kenan Jiang, Xuehai He, Ruize Xu, and Xin~Eric Wang.
\newblock {ComCLIP:} training-free compositional image and text matching.
\newblock In \emph{arXiv:2211.13854}, 2024.

\bibitem[Kamath et~al.(2023)Kamath, Hessel, and Chang]{kamath2023text}
Amita Kamath, Jack Hessel, and Kai-Wei Chang.
\newblock Text encoders bottleneck compositionality in contrastive vision-language models.
\newblock In \emph{Conference on Empirical Methods in Natural Language Processing (EMNLP)}, 2023.

\bibitem[Kenton \& Toutanova(2019)Kenton and Toutanova]{devlin-etal-2019-bert}
Jacob Devlin Ming-Wei~Chang Kenton and Lee~Kristina Toutanova.
\newblock {BERT}: Pre-training of deep bidirectional transformers for language understanding.
\newblock In \emph{Proceedings of NAACL}, 2019.

\bibitem[Kirillov et~al.(2023)Kirillov, Mintun, Ravi, Mao, Rolland, Gustafson, Xiao, Whitehead, Berg, Lo, Doll{\'a}r, and Girshick]{kirillov2023segany}
Alexander Kirillov, Eric Mintun, Nikhila Ravi, Hanzi Mao, Chloe Rolland, Laura Gustafson, Tete Xiao, Spencer Whitehead, Alexander~C. Berg, Wan-Yen Lo, Piotr Doll{\'a}r, and Ross Girshick.
\newblock Segment anything.
\newblock In \emph{arXiv:2304.02643}, 2023.

\bibitem[Krishna et~al.(2017)Krishna, Zhu, Groth, Johnson, Hata, Kravitz, Chen, Kalantidis, Li, Shamma, et~al.]{visual_genome}
Ranjay Krishna, Yuke Zhu, Oliver Groth, Justin Johnson, Kenji Hata, Joshua Kravitz, Stephanie Chen, Yannis Kalantidis, Li-Jia Li, David~A Shamma, et~al.
\newblock Visual genome: Connecting language and vision using crowdsourced dense image annotations.
\newblock \emph{International Journal of Computer Vision}, 123:\penalty0 32--73, 2017.

\bibitem[Krojer et~al.(2023)Krojer, Poole-Dayan, Voleti, Pal, and Reddy]{krojer2023diffusionmodelsvisionandlanguagereasoners}
Benno Krojer, Elinor Poole-Dayan, Vikram Voleti, Christopher Pal, and Siva Reddy.
\newblock Are diffusion models vision-and-language reasoners?
\newblock In \emph{NeurIPS}, 2023.

\bibitem[Leviathan et~al.(2024)Leviathan, Kalman, and Matias]{selectiveattention}
Yaniv Leviathan, Matan Kalman, and Yossi Matias.
\newblock Selective attention improves transformer.
\newblock In \emph{arXiv:2410.02703}, 2024.

\bibitem[Li et~al.(2023{\natexlab{a}})Li, Prabhudesai, Duggal, Brown, and Pathak]{li2023diffusionmodelsecretlyzeroshot}
Alexander~C. Li, Mihir Prabhudesai, Shivam Duggal, Ellis Brown, and Deepak Pathak.
\newblock Your diffusion model is secretly a zero-shot classifier.
\newblock In \emph{ICCV}, 2023{\natexlab{a}}.

\bibitem[Li et~al.(2022)Li, Li, Xiong, and Hoi]{blip}
Junnan Li, Dongxu Li, Caiming Xiong, and Steven Hoi.
\newblock Blip: Bootstrapping language-image pre-training for unified vision-language understanding and generation.
\newblock In \emph{ICML}, 2022.

\bibitem[Li et~al.(2023{\natexlab{b}})Li, Li, Savarese, and Hoi]{BLIP-2}
Junnan Li, Dongxu Li, Silvio Savarese, and Steven Hoi.
\newblock {BLIP-2: B}ootstrapping language-image pre-training with frozen image encoders and large language models.
\newblock In \emph{ICML}, 2023{\natexlab{b}}.

\bibitem[Li et~al.(2023{\natexlab{c}})Li, Li, Savarese, and Hoi]{blip2}
Junnan Li, Dongxu Li, Silvio Savarese, and Steven Hoi.
\newblock Blip-2: Bootstrapping language-image pre-training with frozen image encoders and large language models.
\newblock In \emph{ICML}, 2023{\natexlab{c}}.

\bibitem[Li et~al.(2024{\natexlab{a}})Li, Chen, Hong, Chen, Chen, Shen, and Gan]{li2024covlm}
Junyan Li, Delin Chen, Yining Hong, Zhenfang Chen, Peihao Chen, Yikang Shen, and Chuang Gan.
\newblock Co{VLM}: Composing visual entities and relationships in large language models via communicative decoding.
\newblock In \emph{ICLR}, 2024{\natexlab{a}}.

\bibitem[Li et~al.(2024{\natexlab{b}})Li, Wu, and He]{li2024learningcorrectionefficienttuning}
Rongjie Li, Yu~Wu, and Xuming He.
\newblock Learning by correction: Efficient tuning task for zero-shot generative vision-language reasoning.
\newblock In \emph{CVPR}, 2024{\natexlab{b}}.

\bibitem[Li et~al.(2024{\natexlab{c}})Li, Tu, Hui, Wang, Zhao, Xiao, Ren, Mei, Liu, Zheng, Zhou, and Xie]{li2024recaptionbillionswebimages}
Xianhang Li, Haoqin Tu, Mude Hui, Zeyu Wang, Bingchen Zhao, Junfei Xiao, Sucheng Ren, Jieru Mei, Qing Liu, Huangjie Zheng, Yuyin Zhou, and Cihang Xie.
\newblock What if we recaption billions of web images with llama-3?
\newblock In \emph{arXiv:2406.08478}, 2024{\natexlab{c}}.

\bibitem[Li et~al.(2019)Li, Xu, Liu, Huang, Xu, Chen, Ma, Wang, Fang, and Lu]{hake}
Yong-Lu Li, Liang Xu, Xinpeng Liu, Xijie Huang, Yue Xu, Mingyang Chen, Ze~Ma, Shiyi Wang, Hao-Shu Fang, and Cewu Lu.
\newblock Hake: Human activity knowledge engine.
\newblock In \emph{arXiv:1904.06539}, 2019.

\bibitem[Lin et~al.(2014)Lin, Maire, Belongie, Hays, Perona, Ramanan, Doll{\'a}r, and Zitnick]{coco}
Tsung-Yi Lin, Michael Maire, Serge Belongie, James Hays, Pietro Perona, Deva Ramanan, Piotr Doll{\'a}r, and C~Lawrence Zitnick.
\newblock {Microsoft COCO: Common objects in context}.
\newblock In \emph{ECCV}, 2014.

\bibitem[Lin et~al.(2024)Lin, Chen, Pathak, Zhang, and Ramanan]{Revisiting}
Zhiqiu Lin, Xinyue Chen, Deepak Pathak, Pengchuan Zhang, and Deva Ramanan.
\newblock Revisiting the role of language priors in vision-language models.
\newblock In \emph{ICML}, 2024.

\bibitem[Lin et~al.(2025)Lin, Pathak, Li, Li, Xia, Neubig, Zhang, and Ramanan]{instructblip-vqa-scores}
Zhiqiu Lin, Deepak Pathak, Baiqi Li, Jiayao Li, Xide Xia, Graham Neubig, Pengchuan Zhang, and Deva Ramanan.
\newblock Evaluating text-to-visual generation with image-to-text generation.
\newblock In \emph{ECCV}, 2025.

\bibitem[Liu et~al.(2023)Liu, Li, Wu, and Lee]{LLaVA}
Haotian Liu, Chunyuan Li, Qingyang Wu, and Yong~Jae Lee.
\newblock Visual instruction tuning.
\newblock In \emph{NeurIPS}, 2023.

\bibitem[Liu et~al.(2024)Liu, Li, Li, Li, Zhang, Shen, and Lee]{llavanext}
Haotian Liu, Chunyuan Li, Yuheng Li, Bo~Li, Yuanhan Zhang, Sheng Shen, and Yong~Jae Lee.
\newblock Llava-next: Improved reasoning, ocr, and world knowledge, 2024.
\newblock URL \url{https://llava-vl.github.io/blog/2024-01-30-llava-next/}.

\bibitem[Ma et~al.(2023)Ma, Hong, Gul, Gandhi, Gao, and Krishna]{crepe}
Zixian Ma, Jerry Hong, Mustafa~Omer Gul, Mona Gandhi, Irena Gao, and Ranjay Krishna.
\newblock {CREPE: Can vision-language foundation models reason compositionally?}
\newblock In \emph{CVPR}, 2023.

\bibitem[Marcus et~al.(1993)Marcus, Santorini, and Marcinkiewicz]{parser-ranking}
Mitchell~P. Marcus, Beatrice Santorini, and Mary~Ann Marcinkiewicz.
\newblock Building a large annotated corpus of {E}nglish: The {P}enn {T}reebank.
\newblock \emph{Computational Linguistics}, 19\penalty0 (2):\penalty0 313--330, 1993.

\bibitem[Momeni et~al.(2023)Momeni, Caron, Nagrani, Zisserman, and Schmid]{momeni2023verbsactionimprovingverb}
Liliane Momeni, Mathilde Caron, Arsha Nagrani, Andrew Zisserman, and Cordelia Schmid.
\newblock Verbs in action: Improving verb understanding in video-language models.
\newblock In \emph{arXiv:2304.06708}, 2023.

\bibitem[Mrini et~al.(2019)Mrini, Dernoncourt, Tran, Bui, Chang, and Nakashole]{best_parser}
Khalil Mrini, Franck Dernoncourt, Quan Tran, Trung Bui, Walter Chang, and Ndapa Nakashole.
\newblock Rethinking self-attention: Towards interpretability in neural parsing.
\newblock In \emph{arXiv:1911.03875}, 2019.

\bibitem[Nivre(2005)]{Nivre2005DependencyGA}
Joakim Nivre.
\newblock Dependency grammar and dependency parsing, 2005.
\newblock URL \url{https://api.semanticscholar.org/CorpusID:16318436}.

\bibitem[Obr{\^e}bski \& Gralinski(2004)Obr{\^e}bski and Gralinski]{Obrbski2004SomeNO}
Tomasz Obr{\^e}bski and Filip Gralinski.
\newblock Some notes on generative capacity of dependency grammar.
\newblock In \emph{Workshop On Recent Advances In Dependency Grammar}, 2004.

\bibitem[Oh et~al.(2024)Oh, Cho, Kim, Kweon, and Kim]{oh2024preserving}
Youngtaek Oh, Jae~Won Cho, Dong-Jin Kim, In~So Kweon, and Junmo Kim.
\newblock Preserving multi-modal capabilities of pre-trained {VLMs} for improving vision-linguistic compositionality.
\newblock In \emph{arXiv:2410.05210}, 2024.

\bibitem[OpenAI(2023)]{gpt4v}
OpenAI.
\newblock Gpt-4v(ision) system card, 2023.
\newblock URL \url{https://api.semanticscholar.org/CorpusID:263218031}.

\bibitem[Parascandolo et~al.(2018)Parascandolo, Rojas{-}Carulla, Kilbertus, and Sch{\"{o}}lkopf]{DBLP:journals/corr/abs-1712-00961}
Giambattista Parascandolo, Mateo Rojas{-}Carulla, Niki Kilbertus, and Bernhard Sch{\"{o}}lkopf.
\newblock Learning independent causal mechanisms.
\newblock In \emph{ICML}, 2018.

\bibitem[Pearl(2009)]{CausalInference}
Judea Pearl.
\newblock Causal inference in statistics: An overview.
\newblock \emph{Statistics Surveys}, 3:\penalty0 96--146, 2009.

\bibitem[Pearl \& Verma(1995)Pearl and Verma]{PEARL1995789}
Judea Pearl and Thomas~S. Verma.
\newblock A theory of inferred causation.
\newblock In \emph{Logic, Methodology and Philosophy of Science IX}, volume 134, pp.\  789--811. Elsevier, 1995.

\bibitem[Peng et~al.(2024)Peng, Xie, You, Lan, and Wu]{clip-spec}
Wujian Peng, Sicheng Xie, Zuyao You, Shiyi Lan, and Zuxuan Wu.
\newblock Synthesize, diagnose, and optimize: Towards fine-grained vision-language understanding.
\newblock In \emph{CVPR}, 2024.

\bibitem[Perry et~al.(2022)Perry, von K\"{u}gelgen, and Sch\"{o}lkopf]{Causaldiscovery}
Ronan Perry, Julius von K\"{u}gelgen, and Bernhard Sch\"{o}lkopf.
\newblock Causal discovery in heterogeneous environments under the sparse mechanism shift hypothesis.
\newblock In \emph{NeurIPS}, 2022.

\bibitem[Pham et~al.(2021)Pham, Kafle, Lin, Ding, Cohen, Tran, and Shrivastava]{vaw}
Khoi Pham, Kushal Kafle, Zhe Lin, Zhihong Ding, Scott Cohen, Quan Tran, and Abhinav Shrivastava.
\newblock Learning to predict visual attributes in the wild.
\newblock In \emph{CVPR}, 2021.

\bibitem[Pratt et~al.(2020)Pratt, Yatskar, Weihs, Farhadi, and Kembhavi]{swig}
Sarah Pratt, Mark Yatskar, Luca Weihs, Ali Farhadi, and Aniruddha Kembhavi.
\newblock Grounded situation recognition.
\newblock In \emph{ECCV}, 2020.

\bibitem[Radford et~al.(2021)Radford, Kim, Hallacy, Ramesh, Goh, Agarwal, Sastry, Askell, Mishkin, Clark, et~al.]{radford2021learning}
Alec Radford, Jong~Wook Kim, Chris Hallacy, Aditya Ramesh, Gabriel Goh, Sandhini Agarwal, Girish Sastry, Amanda Askell, Pamela Mishkin, Jack Clark, et~al.
\newblock Learning transferable visual models from natural language supervision.
\newblock In \emph{ICML}, 2021.

\bibitem[Rombach et~al.(2022)Rombach, Blattmann, Lorenz, Esser, and Ommer]{stableDiffusion}
Robin Rombach, Andreas Blattmann, Dominik Lorenz, Patrick Esser, and Bj{\"{o}}rn Ommer.
\newblock High-resolution image synthesis with latent diffusion models.
\newblock In \emph{CVPR}, 2022.

\bibitem[Sahin et~al.(2024)Sahin, Li, Khan, Cremers, and Tresp]{sahin2023enhancingmultimodalcompositionalreasoning}
Ugur Sahin, Hang Li, Qadeer Khan, Daniel Cremers, and Volker Tresp.
\newblock Enhancing multimodal compositional reasoning of visual language models with generative negative mining.
\newblock In \emph{WACV}, 2024.

\bibitem[Sch{\"o}lkopf et~al.(2021)Sch{\"o}lkopf, Locatello, Bauer, Ke, Kalchbrenner, Goyal, and Bengio]{TowardCausal}
Bernhard Sch{\"o}lkopf, Francesco Locatello, Stefan Bauer, Nan~Rosemary Ke, Nal Kalchbrenner, Anirudh Goyal, and Yoshua Bengio.
\newblock Toward causal representation learning.
\newblock \emph{Proceedings of the IEEE}, 109:\penalty0 612--634, 2021.

\bibitem[Sharma et~al.(2018)Sharma, Ding, Goodman, and Soricut]{cc3m}
Piyush Sharma, Nan Ding, Sebastian Goodman, and Radu Soricut.
\newblock Conceptual captions: A cleaned, hypernymed, image alt-text dataset for automatic image captioning.
\newblock In \emph{ACL}, 2018.

\bibitem[Silveira et~al.(2014)Silveira, Dozat, de~Marneffe, Bowman, Connor, Bauer, and Manning]{dependency_relations}
Natalia Silveira, Timothy Dozat, Marie-Catherine de~Marneffe, Samuel Bowman, Miriam Connor, John Bauer, and Chris Manning.
\newblock A gold standard dependency corpus for {E}nglish.
\newblock In \emph{Proceedings of the Ninth International Conference on Language Resources and Evaluation ({LREC})}, 2014.

\bibitem[Singh et~al.(2021)Singh, Hu, Goswami, Couairon, Galuba, Rohrbach, and Kiela]{FLAVA}
Amanpreet Singh, Ronghang Hu, Vedanuj Goswami, Guillaume Couairon, Wojciech Galuba, Marcus Rohrbach, and Douwe Kiela.
\newblock {FLAVA: A} foundational language and vision alignment model.
\newblock \emph{CVPR}, 2021.

\bibitem[Singh et~al.(2024)Singh, Khan, and Tapaswi]{s2024figclipfinegrainedclipadaptation}
Darshan Singh, Zeeshan Khan, and Makarand Tapaswi.
\newblock {FiGCLIP: Fine-Grained CLIP Adaptation via Densely Annotated Videos}.
\newblock In \emph{arXiv:2401.07669}, 2024.

\bibitem[Singh et~al.(2023)Singh, Zhang, Wang, Wang, Xiong, Du, and Chen]{singh-etal-2023-coarse}
Harman Singh, Pengchuan Zhang, Qifan Wang, Mengjiao Wang, Wenhan Xiong, Jingfei Du, and Yu~Chen.
\newblock Coarse-to-fine contrastive learning in image-text-graph space for improved vision-language compositionality.
\newblock In \emph{Conference on Empirical Methods in Natural Language Processing (EMNLP)}, 2023.

\bibitem[Song et~al.(2022)Song, Dong, Zhang, Liu, and Wei]{song2022clipmodelsfewshotlearners}
Haoyu Song, Li~Dong, Wei-Nan Zhang, Ting Liu, and Furu Wei.
\newblock {CLIP M}odels are few-shot learners: Empirical studies on {VQA} and visual entailment.
\newblock In \emph{ACL}, 2022.

\bibitem[Tenenbaum et~al.(2007)Tenenbaum, Griffiths, and Niyogi]{Intuitive}
Joshua~B. Tenenbaum, Thomas~L. Griffiths, and Sourabh Niyogi.
\newblock Intuitive theories as grammars for causal inference.
\newblock In \emph{Causal Learning: Psychology, Philosophy, and Computation}. Oxford University Press, 2007.

\bibitem[Thrush et~al.(2022)Thrush, Jiang, Bartolo, Singh, Williams, Kiela, and Ross]{Winoground}
Tristan Thrush, Ryan Jiang, Max Bartolo, Amanpreet Singh, Adina Williams, Douwe Kiela, and Candace Ross.
\newblock Winoground: Probing vision and language models for visio-linguistic compositionality.
\newblock In \emph{CVPR}, 2022.

\bibitem[Tong et~al.(2024)Tong, Liu, Zhai, Ma, LeCun, and Xie]{eyes-wide-shut}
Shengbang Tong, Zhuang Liu, Yuexiang Zhai, Yi~Ma, Yann LeCun, and Saining Xie.
\newblock Eyes wide shut? exploring the visual shortcomings of multimodal llms.
\newblock In \emph{CVPR}, 2024.

\bibitem[Tschannen et~al.(2023)Tschannen, Kumar, Steiner, Zhai, Houlsby, and Beyer]{Cap}
Michael Tschannen, Manoj Kumar, Andreas~Peter Steiner, Xiaohua Zhai, Neil Houlsby, and Lucas Beyer.
\newblock Image captioners are scalable vision learners too.
\newblock In \emph{NeurIPS}, 2023.

\bibitem[Vaswani et~al.(2017)Vaswani, Shazeer, Parmar, Uszkoreit, Jones, Gomez, Kaiser, and Polosukhin]{attention-is-all-you-need}
Ashish Vaswani, Noam Shazeer, Niki Parmar, Jakob Uszkoreit, Llion Jones, Aidan~N. Gomez, Lukasz Kaiser, and Illia Polosukhin.
\newblock Attention is all you need.
\newblock In \emph{NeurIPS}, 2017.

\bibitem[Wan et~al.(2024)Wan, Cho, Stengel-Eskin, and Bansal]{CRG}
David Wan, Jaemin Cho, Elias Stengel-Eskin, and Mohit Bansal.
\newblock Contrastive region guidance: Improving grounding in vision-language models without training.
\newblock In \emph{arXiv:2403.02325}, 2024.

\bibitem[Wazni et~al.(2024)Wazni, Lo, and Sadrzadeh]{wazni-etal-2024-verbclip}
Hadi Wazni, Kin~Ian Lo, and Mehrnoosh Sadrzadeh.
\newblock {V}erb{CLIP}: Improving verb understanding in vision-language models with compositional structures.
\newblock In \emph{Proceedings of the 3rd Workshop on Advances in Language and Vision Research (ALVR)}, 2024.

\bibitem[Wysoczańska et~al.(2024)Wysoczańska, Siméoni, Ramamonjisoa, Bursuc, Trzciński, and Pérez]{clipdinoiserteachingclipdino}
Monika Wysoczańska, Oriane Siméoni, Michaël Ramamonjisoa, Andrei Bursuc, Tomasz Trzciński, and Patrick Pérez.
\newblock {CLIP-DINOiser: Teaching CLIP a few DINO tricks for open-vocabulary semantic segmentation}.
\newblock In \emph{arXiv:2312.12359}, 2024.

\bibitem[Yang \& Wan(2022)Yang and Wan]{yang2022dependencybasedmixturelanguagemodels}
Zhixian Yang and Xiaojun Wan.
\newblock Dependency-based mixture language models.
\newblock In \emph{ACL}, 2022.

\bibitem[Ye et~al.(2024)Ye, Dong, Xia, Sun, Zhu, Huang, and Wei]{differentialtransformer}
Tianzhu Ye, Li~Dong, Yuqing Xia, Yutao Sun, Yi~Zhu, Gao Huang, and Furu Wei.
\newblock Differential transformer.
\newblock In \emph{arXiv:2410.05258}, 2024.

\bibitem[Yellinek et~al.(2023)Yellinek, Karlinsky, and Giryes]{3VL}
Nir Yellinek, Leonid Karlinsky, and Raja Giryes.
\newblock {3VL:} using trees to teach vision \& language models compositional concepts.
\newblock In \emph{arXiv:2312.17345}, 2023.

\bibitem[Yuksekgonul et~al.(2023)Yuksekgonul, Bianchi, Kalluri, Jurafsky, and Zou]{NegCLIP}
Mert Yuksekgonul, Federico Bianchi, Pratyusha Kalluri, Dan Jurafsky, and James~Y. Zou.
\newblock When and why vision-language models behave like bags-of-words, and what to do about it?
\newblock In \emph{ICLR}, 2023.

\bibitem[Zeng et~al.(2022)Zeng, Zhang, and Li]{xvlm}
Yan Zeng, Xinsong Zhang, and Hang Li.
\newblock Multi-grained vision language pre-training: Aligning texts with visual concepts.
\newblock In \emph{ICML}, 2022.

\bibitem[Zhang et~al.(2024)Zhang, Awal, and Agrawal]{zhang2024contrastingintramodalrankingcrossmodal}
Le~Zhang, Rabiul Awal, and Aishwarya Agrawal.
\newblock Contrasting intra-modal and ranking cross-modal hard negatives to enhance visio-linguistic compositional understanding.
\newblock In \emph{CVPR}, 2024.

\bibitem[Zhang et~al.(2022)Zhang, Lijie, Xiao, and Wu]{SynCLM}
Shuai Zhang, Wang Lijie, Xinyan Xiao, and Hua Wu.
\newblock Syntax-guided contrastive learning for pre-trained language model.
\newblock In \emph{Findings of the Association for Computational Linguistics}, 2022.

\bibitem[Zhang et~al.(2020)Zhang, Li, and Zhang]{zhang-etal-2020-efficient}
Yu~Zhang, Zhenghua Li, and Min Zhang.
\newblock Efficient second-order {T}ree{CRF} for neural dependency parsing.
\newblock In \emph{ACL}, 2020.

\bibitem[Zhao et~al.(2022)Zhao, Zhang, Zhu, Shen, Lee, Lu, and Yin]{vl_checklist}
Tiancheng Zhao, Tianqi Zhang, Mingwei Zhu, Haozhan Shen, Kyusong Lee, Xiaopeng Lu, and Jianwei Yin.
\newblock Vl-checklist: Evaluating pre-trained vision-language models with objects, attributes and relations.
\newblock In \emph{arXiv:2207.00221}, 2022.

\bibitem[Zheng et~al.(2024)Zheng, Zhang, Kembhavi, and Krishna]{zheng2024iteratedlearningimprovescompositionality}
Chenhao Zheng, Jieyu Zhang, Aniruddha Kembhavi, and Ranjay Krishna.
\newblock Iterated learning improves compositionality in large vision-language models.
\newblock In \emph{CVPR}, 2024.

\bibitem[Zhu et~al.(2024)Zhu, Chen, Shen, Li, and Elhoseiny]{minigpt4}
Deyao Zhu, Jun Chen, Xiaoqian Shen, Xiang Li, and Mohamed Elhoseiny.
\newblock Mini{GPT}-4: Enhancing vision-language understanding with advanced large language models.
\newblock In \emph{ICLR}, 2024.

\end{thebibliography}
\bibliographystyle{iclr2025_conference}
\newpage
\appendix

\section{Causal Graphical Models}
\label{app.Factorization}

A Causal Graphical Model (CGM)  over $n$ random variables $\mathbf{X} = \{ X_1, ..., X_n \}$
is defined \citep{Causaldiscovery} as ${\cal M}(G, \mathbb{P}_{\mathbf{X}})$, where:
(1) $G$ is a  directed acyclic graph with vertices $\mathbf{X}$ and edges $X_i \rightarrow X_j$ iff $X_i$ is a direct cause of $X_j$;
(2) $\mathbb{P}_{\mathbf{X}}$ is the joint distribution of $\mathbf{X}$ which follows the {\em disentangled (or causal) factorization}
\citep{TowardCausal,Causaldiscovery}:

\begin{equation}
\label{eq.disentangled-factorization}
    P(X_1, ..., X_n ) = \prod_{j=1}^n P(X_j | \mathbf{PA}(X_j)),
\end{equation}

\noindent
where $\mathbf{PA}(X_j)$ is the set of {\em parents} (direct causes) of $X_j$ in $G$.
The difference between a CGM and a Directed Graphical Model is that the former assumes that $\mathbf{PA}(X_j)$ are direct causes of $X_j$. Although a formal proof that a statistical dependence  is also a causal relation using only observational data is notoriously difficult \citep{CausalInference}, in this paper we assume that the linguistic dependencies \citep{Nivre2005DependencyGA} extracted by a dependency parser   have a causal nature because they represent a strict linguistic association between the ``head'' and its ``dependent''. 
Specifically, Dependency Grammars (\Cref{sec.RelatedWork}) can be considered as (probabilistic) generative grammars
\citep{Nivre2005DependencyGA,chen-manning-2014-fast,Obrbski2004SomeNO,Diaconescu}, in which a dependence  between a ``head'' word and its ``dependent'' word can be extracted using context-free generative rules.
We interpret these rules as causal {\em mechanisms} \citep{TowardCausal}, which describe the causal influence of  generating a specific 
``dependent'' word given the value of the `head'' word. 
We leave as future work the possibility of replacing the dependency grammars used in this paper with other grammars such as, for instance, the causal grammars proposed in \citep{Intuitive}, as well as the possible introduction of counterfactual reasoning \citep{CausalInference} in our framework.

In \Cref{sec.Method},
\Cref{eq.jointdistrib-disentangled} is obtained using \Cref{eq.disentangled-factorization},  the definition of conditional distribution and the assumption that 
$S_1, ..., S_n$ and $Z_1, ..., Z_m$ are independent of each other.
Finally, the cardinality of  $\{ W_1, ..., W_{j-1} \}$ in \Cref{eq.jointdistrib-AR} is, on average, $\frac{n}{2}$, while, assuming a balanced DT $\mathcal{T}$, the cardinality of   
$\{ W_{i_1}, ...,  W_{i_k} \} \subset \mathbf{PA}(W_j)$ is, on average,  $O(\log(n))$. The consequence of this is that
the conditional distributions learned at training time (\Cref{eq.jointdistrib-disentangled})
and used at inference time (\Cref{eq.log-likelihood})
are sparser than those learned by a standard AR model (\Cref{sec.Method}).

\section{The FG-OVD Dataset}
\label{app.fg_ovd}

In this section, we describe the  compositional benchmark based on the FG-OVD dataset which we propose in this paper and which was used in \Cref{sec.Experiments}. The FG-OVD dataset was originally proposed to evaluate the fine-grained discriminative capabilities of open-vocabulary detectors in object detection tasks. 
Each image usually contains multiple objects, where each object is associated with both a bounding box and a corresponding caption.
We use the bounding box to crop the image and the corresponding caption as the true caption. The 
cropped images are resized to a resolution of 
$224 \times 224$.
Then, each object image is associated with several false captions (on average, ten), selected based on the original FG-OVD Trivial, Easy, Medium, and Hard tasks \citep{FG-OVD}.
Specifically,
in the Trivial task,  negative captions are randomly sampled from unrelated objects (of different images), offering a basic challenge for retrieval. The Easy, Medium, and Hard tasks progressively increase the difficulty by generating negative captions starting from the true caption.
For instance, in the Easy task, three attributes of the true caption are replaced with three unrelated attributes. In the Medium and the Hard task, two and one attributes are replaced, respectively. The rationale is that the less the true sentence is modified, the harder is to distinguish the false from the true captions \citep{FG-OVD}. 
Specifically, as fewer attributes are replaced, the distinction between correct and incorrect captions becomes more subtle, requiring the model to capture increasingly fine-grained, compositional details in the image-text pairs.
 \Cref{tab:datasets_summary} reports the number of testing images for each task, while different examples are shown in \Cref{subsec.qualitative_results}.


\section{Additional experiments}
\label{app.AdditionalExp}

\subsection{Additional ablations}
\label{subsec.additional_ablations} 

In \Cref{tab.ablation} we show  an extension of \Cref{tab.Ablation-all-the-rest} containing all the possible combinations of the components analyzed in \Cref{tab.Ablation-all-the-rest}.
 For instance, this table shows the importance of using two visual feature layers in some of the datasets. Indeed, when both layers are used, the model can leverage not only high-level, abstract visual features but also more detailed, lower-level information. This is particularly important in tasks which require attention to object details, where the additional layer helps to capture a more nuanced representation of the input. For instance, in datasets like ColorSwap, this deeper feature extraction leads to drastic improvements (observed across all parsers), which is probably due to the better representation of the color/texture appearance in the lower level features. 
 On the other hand, the use of mask-specific tokens also plays a significant role. Although the improvement magnitude  varies depending on the dataset and the task, the overall trend indicates that using mask-specific tokens contributes positively to the accuracy.

 Comparing the results of COGT-CLIP with COGT-CLIP+ and COGT-XVLM with COGT-XVLM+ (\Cref{tab.SOTA-comparison} and \Cref{tab.SOTA-comparison-xvlm}), 
the mean improvement of the larger-training versions with respect to the COCO-only trained models is about 9 points averaged across all the datasets. This shows that COGT can benefit from larger training and, indirectly, that the noisier captions in CC3M can be effectively parsed by our parser. 
Finally,  a recent trend in VLM fine-tuning and/or  pre-training adopts LLMs to create or re-write the textual annotations \citep{li2024recaptionbillionswebimages,DAC}, which can in principle help the dependency parser with very noisy captions. We leave this as future work.

\begin{table*}[t]
    \caption{Extended version of  \Cref{tab.Ablation-all-the-rest}}
    \vspace{0.3em}
    \label{tab.ablation}
    \centering
    \setlength{\tabcolsep}{0.05em}
    \resizebox{\textwidth}{!}{%
    \begin{tabular}{lccc c ccc c cccc c cccc c c c c c}
    \toprule
    & & & & \multicolumn{3}{c}{\textbf{ARO}} & & \multicolumn{4}{c}{\textbf{SugarCrepe}} & & \multicolumn{4}{c}{\textbf{VL-Checklist}}  & & \textbf{ColorSwap} & & \textbf{FG-OVD} \\
    \cmidrule{5-8} \cmidrule{9-12} \cmidrule{14-17} \cmidrule{19-19} \cmidrule{21-21}  
     \textbf{Parser} & \textbf{Layers} & \textbf{Mask Specific} & & \textbf{Relation} & \textbf{Attribute} & \textbf{Avg} & & \textbf{Add} & \textbf{Replace} & \textbf{Swap} & \textbf{Avg} & & \textbf{Attribute} & \textbf{Object} & \textbf{Relation} & \textbf{Avg} & & \textbf{ITT}& & \textbf{Avg} & & \textbf{Avg} \\
    \midrule
    Deep Biaffine & 1 & \xmark & & 87.19 & 84.92 & 86.06 & & 98.31 & 85.52 & 78.56 & 87.46 & & 87.28 & 78.56 & 90.24 & 85.36 & & 14.00 & & 43.91 & & 63.36 \\
    Deep Biaffine & 1 & \cmark & & 87.34 & 86.93 & 87.14 & & 98.02 & 85.17 & 80.77 & 87.99 & & 86.47 & 79.29 & 89.41 & 85.05 & & 38.00 & & 44.77 & & 68.59 \\
    Deep Biaffine & 2 & \xmark & & 86.10 & 86.51 & 86.31 & & 98.48 & 85.37 & 80.59 & 88.14 & & 87.60 & 77.97 & 91.51 & 85.69 & & 60.33 & & 46.46 & & 73.39 \\
    Deep Biaffine & 2 & \cmark & & 86.56 & 89.10 & 87.83 & & 98.11 & 85.80 & 81.49 & 88.46 & & 87.02 & 78.30 & 87.75 & 84.35 & & 61.33 & & 44.74 & & 73.34 \\
    \midrule
    CRFPar & 1 & \xmark & & 86.46 & 84.36 & 87.83 & & 98.45 & 84.15 & 78.65 & 87.08 & & 87.51 & 77.04 & 90.15 & 84.90 & & 14.00 & & 45.83 & & 63.93 \\
    CRFPar & 1 & \cmark & & 86.85 & 88.33 & 87.59 & & 98.10 & 86.00 & 81.34 & 88.47 & & 86.80 & 78.49 & 88.80 & 84.70 & & 42.33 & & 42.27 & & 69.07 \\
    CRFPar & 2 & \xmark & & 85.53 & 86.67 & 86.10 & & 98.74 & 85.58 & 81.84 & 88.72 & & 87.43 & 77.33 & 90.30 & 85.02 & & 59.33 & & 44.08 & & 72.65 \\
    CRFPar & 2 & \cmark & & 85.68 & 88.34 & 87.01 & & 98.16 & 84.94 & 80.30 & 87.80 & & 86.99 & 77.68 & 87.09 & 83.92 & & 56.33 & & 43.74 & & 71.76 \\
    \midrule
    Deep Biaffine + RoBERTa & 1 & \xmark & & 86.82 & 86.76 & 86.79 & & 98.69 & 86.59 & 80.35 & 88.54 & & 85.31 & 79.11 & 89.88 & 84.76 & & 28.33 & & 43.92 & & 66.47 \\
    Deep Biaffine + RoBERTa & 1 & \cmark & & 86.82 & 89.67 & 88.25 & & 98.26 & 86.56 & 82.33 & 89.05 & & 84.41 & 78.94 & 89.54 & 84.30 & & 45.00 & & 45.63 & & 70.45 \\
    Deep Biaffine + RoBERTa & 2 & \xmark & & 84.75 & 86.16 & 85.46 & & 98.86 & 84.37 & 80.25 & 87.82 & & 83.79 & 78.24 & 90.84 & 84.29 & & 58.00 & & 46.99 & & 72.51 \\
    Deep Biaffine + RoBERTa & 2 & \cmark & & 87.56 & 90.26 & 88.91 & & 98.26 & 87.10 & 83.14 & 89.50 & & 86.07 & 78.91 & 89.37 & 84.78 & & 61.33 & & 51.48 & & 75.20 \\
    \bottomrule
\end{tabular}%
}
\end{table*}

\subsection{Winoground and the Multimodal Visual Pattern (MMVP) benchmarks}
\label{subsec.winoground_eyes_wide_shut} 

In \Cref{tab.winoground_mmvp} we show additional results obtained using the Winoground~\citep{Winoground} and the MMVP~\citep{eyes-wide-shut} benchmarks. 
MMVP is a relatively small dataset, with 9 tasks but only  15 samples per task (135 total  samples), thus performance measured on this benchmark has a limited statistical significance.
Nevertheless, also in this dataset
\ours-InstructBLIP+  largely outperforms the second best tested model (DAC-LLM) by 4.45 points. On the other hand, on Winoground,  \ours-InstructBLIP+,
despite an improvement of  +7.5 points with respect to InstructBLIP zero-shot,
it is largely outperformed by InstructBLIP-VQA Score~\citep{instructblip-vqa-scores}.
InstructBLIP-VQA could eventually be merged with \ours, but we leave this as a future work.

\begin{table*}[t]
    \caption{Model comparison on MMVP and Winoground.}
    \vspace{0.3em}
    \label{tab.winoground_mmvp}
    \centering
    \setlength{\tabcolsep}{0.05em}
    \resizebox{0.6\textwidth}{!}{%
    \begin{tabular}{l c c c c c c c c}
    \toprule
    & \multicolumn{4}{c}{\textbf{MMVP}} & & \multicolumn{3}{c}{\textbf{Winoground}} \\
    \cmidrule{3-4} \cmidrule{7-9}
    \textbf{Model} & & & \textbf{Group Score} & & & \textbf{Group Score}\\
    \midrule
    CLIP~\citep{radford2021learning} & & & 15.55 & & & 8.00 \\
    NegCLIP~\citep{NegCLIP} & & & 26.67 & & & 8.00 \\
    CE-CLIP~\citep{zhang2024contrastingintramodalrankingcrossmodal} & & & 19.25 & & & 5.00  \\
    DAC-SAM~\citep{DAC} & & & 23.70 & & & 6.25  \\
    DAC-LLM~\citep{DAC} & & & 28.14\rlap{\raisebox{0.5ex}{*}} & & & 3.25 \\
    Plausible Adj. Neg~\citep{buettner2023investigatingroleattributecontext} & & & 21.48 & & & 6.75 \\
    InstructBLIP~\citep{instructblip} & & & 14.81 & & & 4.75  \\
    InstructBLIP-VQA Score~\citep{instructblip-vqa-scores} & & & - & & & \textbf{28.50}  \\
    \rowcolor[gray]{0.9}
    COGT-CLIP & & & 20.74 & & & 6.65 \\
    \rowcolor[gray]{0.9}
    COGT-CLIP+ & & & 26.67 & & & 9.75 \\
    \rowcolor[gray]{0.9}
    COGT-XVLM & & & 17.77 & & & 7.25  \\
    \rowcolor[gray]{0.9}
    COGT-XVLM+ & & & \underline{28.88} & & & 10.50\rlap{\raisebox{0.5ex}{*}} \\
    \rowcolor[gray]{0.9}
    COGT+InstructBLIP & & & \underline{28.88} & & & 7.75 \\
    \rowcolor[gray]{0.9}
    COGT+InstructBLIP+ & & & \textbf{32.59} & & &   \underline{12.25} \\
    \bottomrule
    \end{tabular}%
    }
\end{table*}

\subsection{FG-OVD}
\label{subsec.fg-ovd}

\Cref{tab.fg-ovd} reports the  FG-OVD task-specific accuracy of the methods compared in \Cref{tab.SOTA-comparison}.
 In the Zero-shot category, CLIP ~\citep{radford2021learning} shows a solid performance on all but the Hard task ($21.35$ points accuracy).
 The  same applies to those  methods based on CLIP, such as 
 NegCLIP ~\citep{NegCLIP}, GNM ~\citep{sahin2023enhancingmultimodalcompositionalreasoning}, Plausible Adj. Neg ~\citep{buettner2023investigatingroleattributecontext}, and CE-CLIP ~\citep{zhang2024contrastingintramodalrankingcrossmodal},
 which  show  a noticeable drop in performance as the difficulty increases. Among these models, Plausible Adj. Neg ~\citep{buettner2023investigatingroleattributecontext} stands out with a relatively strong performance, especially on the Medium ($43.13$) and Easy ($45.88$) tasks. 
  On the other hand, \ours-CLIP and \ours-XVLM demonstrate a significant improvement in performance across all difficulty levels, particularly on the Hard task, where \ours-CLIP achieves a score of $33.82$, 
  which is greater than all other methods, including  models trained on larger datasets, such as DAC-SAM and DAC-LLM ~\citep{DAC}. 
  These results highlight the strength of \ours when dealing with fine-grained compositional tasks.
Finally, similarly to the results shown in \Cref{sec.SOTA-exp}, \ours-CLIP+, \ours-XVLM+ and \ours-InstrctBLIP+ further increase the advantage over other approaches.
  
\subsection{Computational efficiency}
\label{subsec.efficiency}

\Cref{tab:training_inference_memory} shows the training and inference times for \ours and CLIP-based models. COGT-CLIP and COGT-CLIP+ require respectively 8 and 72 hours  to train on a {\em single} RTX A5000 GPU with a batch size of 128 using the datasets of \Cref{sec.SOTA-exp}.
For comparison, we use
DAC-SAM and DAC-LLM \citep{DAC}, which are the only models we know with publicly available training times that can be directly compared to COGT-CLIP+, as they are trained on a similar dataset of $\sim$3.3M samples. In particular, both DAC-SAM and DAC-LLM complete their training in 12 hours on {\em six} V100 GPUs with a batch size of 32. However, this training time does not include the computationally intensive dense annotation generation pipeline (\Cref{sec.RelatedWork}), which involves BLIP2 and SAM or GPT-Neo-2.7B (for DAC-SAM and DAC-LLM, rispectively). Similarly, the training times for COGT-CLIP and COGT-CLIP+ do not include the DT generations, which however involves a relatively quick preprocessing step (one DT per caption), taking approximately 3 minutes for COCO and 1.5 hours for the combined CC3M, COCO, and Visual Genome datasets.
Moreover, 
we evaluate the computational costs of \ours and CLIP in terms of memory usage and inference time. COGT-CLIP, COGT-CLIP+, and CLIP require 0.73 GB, 0.88 GB, and 1.16 GB of memory, respectively, where the difference with respect to CLIP is mainly due to the fact that COGT does not use
the CLIP textual encoder. Finally, for both COGT-CLIP and COGT-CLIP+, the  inference times reported in \Cref{tab:training_inference_memory} include an additional 0.01 seconds required for generating the DT of the testing caption using the Deep Biaffine + RoBERTa parser, and COGT-CLIP+ is slower than COGT-CLIP because of its larger (four blocks) decoder.
All times are computed using an RTX A5000 GPU with a batch size of 32.

\begin{table}[ht]
\caption{A training and inference times comparison.}
\label{tab:training_inference_memory}
\centering
\resizebox{0.6\textwidth}{!}{%
\begin{tabular}{lccc}
\toprule
\multicolumn{4}{c}{\textbf{Training}} \\
\midrule
\textbf{Model} & \textbf{Training Time (hrs)} & \textbf{Batch Size} & \textbf{GPU Setup} \\
\midrule
COGT-CLIP       & 8                & 128                 & RTX A5000 (Single GPU) \\
COGT-CLIP+      & 72              & 128                 & RTX A5000 (Single GPU) \\
DAC-SAM         & 12        & 32                  & V100 (Six GPUs)    \\
DAC-LLM         & 12         & 32                  & V100 (Six GPUs) \\
\midrule
\multicolumn{4}{c}{\textbf{Inference}} \\
\midrule
\textbf{Model} & \textbf{Inference Time (s)} & \textbf{Batch Size} & \textbf{GPU Setup} \\
\midrule
COGT-CLIP       & 0.07           & 32                 & RTX A5000 (Single GPU) \\
COGT-CLIP+      & 0.09           & 32                 & RTX A5000 (Single GPU) \\
CLIP            & 0.06           & 32                 & RTX A5000 (Single GPU) \\
\bottomrule
\end{tabular}}
\vspace{1mm}
\end{table}

\begin{table*}[t]
    \caption{FG-OVD: task specific results.}
    \vspace{0.3em}
    \label{tab.fg-ovd}
    \centering
    \setlength{\tabcolsep}{1.5em}
    \resizebox{\textwidth}{!}{%
    \begin{tabular}{lc c cccc c c}
    \toprule
    \textbf{Model} & & \textbf{Hard} & \textbf{Medium} & \textbf{Easy} & \textbf{Trivial} & & \textbf{Avg}  \\
    \midrule
    \multicolumn{1}{c}{\textit{Zero-shot}} \\
    CLIP~\citep{radford2021learning} & & 21.35 & 48.75 & 51.73 & 67.48\rlap{\raisebox{0.5ex}{*}} & & 47.33 \\
    InstructBLIP (FlanT5XL)~\citep{instructblip} & & 22.23 & 33.25 & 31.87 & 19.85 & & 26.80 \\
    \midrule
    \multicolumn{1}{c}{\textit{Training on COCO only}} \\
    NegCLIP~\citep{NegCLIP} & & 18.39 & 36.96 & 41.95 & \underline{69.49} & & 41.69 \\
    GNM~\citep{sahin2023enhancingmultimodalcompositionalreasoning} & & 16.08 & 34.74 & 39.88 & 64.48 & & 38.79 \\
    Plausible Adj. Neg~\citep{buettner2023investigatingroleattributecontext} & & 21.35 & 43.13 & 45.88 & \textbf{69.59} & & 44.98 \\
    CE-CLIP~\citep{zhang2024contrastingintramodalrankingcrossmodal} & & 21.86 & 40.36 & 43.11 & 62.53 & & 41.97 \\
    \rowcolor[gray]{0.9}
    Fully-Parallel & & 25.22 & 47.41 & 54.04 & 40.72 & & 41.84 \\
    \rowcolor[gray]{0.9}
    Mixed & & 30.16 & 51.38 & 56.2 & 43.09 & & 45.21 \\
    \rowcolor[gray]{0.9}
    Sequential-AR & & 30.18 & 54.01 & 57.04 & 43.73 & & 46.24 \\
    \rowcolor[gray]{0.9}
    \ours-CLIP & & 33.82 & 59.30 & 61.35 & 51.43 & & 51.48 \\
    \rowcolor[gray]{0.9}
    \ours-XVLM & & 32.69 & 58.52 & 60.05 & 49.22 & & 50.12 \\
    \rowcolor[gray]{0.9}
    \ours-InstructBLIP & & 33.90 & 59.91 & 61.12 & 50.15 & & 51.26 \\
    \midrule
    \multicolumn{1}{c}{\textit{Training on datasets larger than COCO}} \\
    DAC-SAM~\citep{DAC} & & 26.00 & 48.65 & 53.73 & 65.05 & & 48.36 \\
    DAC-LLM~\citep{DAC} & & 25.29 & 52.36 & 56.89 & 63.89 & & 49.60 \\
    \rowcolor[gray]{0.9}
    \ours-CLIP+ & & \underline{55.40} & \underline{81.50} & \underline{85.29} & 57.65 & & 69.96\rlap{\raisebox{0.5ex}{*}} \\
    \rowcolor[gray]{0.9}
    \ours-XVLM+ & & \textbf{58.78} & \textbf{83.86} & \textbf{87.45} & 66.82 & & \textbf{74.22} \\
    \rowcolor[gray]{0.9}
    \ours-InstructBLIP+ & & 53.59\rlap{\raisebox{0.5ex}{*}} & 81.37\rlap{\raisebox{0.5ex}{*}} & 84.52\rlap{\raisebox{0.5ex}{*}} & 63.36 & & \underline{70.72} \\
    \bottomrule
\end{tabular}%
}
\end{table*}

\section{Implementation details}
\label{app.ImplementationDetails}

\subsection{Architectures}
\label{subsec.architectures}
In \ours-CLIP and in \ours-XVLM we use  ViT-B/32 CLIP ~\citep{open_clip} and the Swin-Transformer of XVLM ~\citep{xvlm} as the visual encoder, respectively.
In both cases, we use both the last and the penultimate layer features of the encoder  (\Cref{sec.Decoder}).
In \ours-InstructBLIP, we use the output of the InstructBLIP Q-Former \citep{instructblip} as the visual encoder. Since InstructBLIP needs a textual description of the task (called ``instruction'' \citep{instructblip}),  \ours-InstructBLIP is trained  using the prompts suggested in \citep{instructblip} for  captioning tasks. At inference time, both
the zero-shot results of InstructBLIP and those of \ours-InstructBLIP are obtained
using the prompt ``Write a description for the photo.''. 
The above considerations apply also to the \ours-X+ models.

Independently of the VLM encoder, the  features ${\cal Z}$ are obtained using a mapping network 
${\cal M}$ on top of the corresponding frozen visual encoder (\Cref{sec.Decoder}). 
Our decoder consists of 3 blocks (respectively, 4 blocks in case of \ours-X+), each composed of a multi-head
 Dependency Guided Attention (\Cref{sec.Decoder})
 and a cross-attention layer. Each attention layer is composed of  8 attention heads, with embedding size equal to 512, while we use 12 heads and embedding size equal to 768 in the \ours-X+ models.
 We apply a dropout rate of 0.1 to the residual connections, the attention weights, and the embeddings. 

The differences in the number of trainable parameters among the different baselines in \Cref{tab.Ablation-generative} are only due to the size of the embedding dictionary. {\em Fully-Parallel} has an embedding dictionary consisting of only one $\texttt{MSK}$ token, resulting in a total of 13 million trainable parameters, of which only 512 are dedicated to represent the $\texttt{MSK}$ token. The total number of trainable parameters for  {\em Mixed}, {\em Sequential-AR}, and \ours  is approximately 64 million (${\cal M}$ included). Among these, {\em Sequential-AR} uses an embedding dictionary that matches the size of the CLIP ViT-B/32 textual encoder, while  {\em Mixed} introduces an additional $\texttt{MSK}$ token for parallel processing. In contrast, \ours employs 45 extra $\texttt{MSK}$ tokens, each representing a specific dependency relation extracted by the parser, resulting in a negligible increase in the total parameter count of only 0.04\%. All the decoders in \Cref{tab.Ablation-generative} are composed of three blocks which differ only in their attention masks, and they all  alternate a textual-token embedding attention layer with   a cross-attention  layer with the features ${\cal Z}$ (extracted from the last  and the penultimate layer of the CLIP visual encoder, see \Cref{sec.Decoder}).

\subsection{Tokenization}
\label{subsec.tokenization}

\begin{figure*}[t]
\centering
{\includegraphics[width=\linewidth]{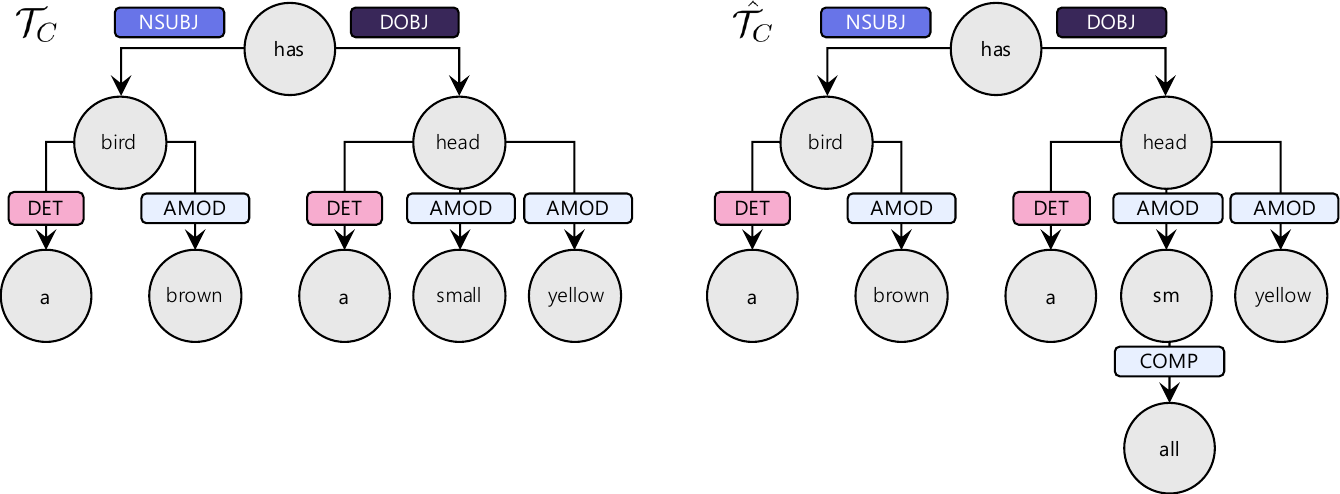}}
\caption{Dependency tree modifications due to the sub-word tokenization. On the left, the original dependency tree $\mathcal{T}$ represents the sentence ``A brown bird with a small yellow head'', and it is the output of a word-level dependency parser, where ``small'' is a single node. On the right, the modified tree, $\hat{\mathcal{T}}$, which accounts for sub-word tokenization by splitting the word 'small' into two nodes: ``sm'' and ``all''. A new syntactic relation, called $\texttt{comp}$, is introduced between these sub-word nodes.}
\label{fig:tokenization}
\end{figure*}

The output of the dependency parser is a tree in which each node  corresponds to a caption word. In contrast, the  \ours decoder uses a standard sub-word tokenization, splitting words into smaller tokens which brings to a smaller embedding dictionary. This discrepancy leads to cases where a word of the dependency tree is split into multiple tokens by the \ours decoder's tokenizer. We modify the dependency tree to handle this mismatch: if a word $w_j$ is split into sub-tokens $w_{j_{1}}$ and $w_{j_{2}}$ by the \ours decoder's tokenizer, then 
we create a new node for $w_{j_{2}}$ and a new edge between $w_{j_{2}}$ and $w_{j_{1}}$ associated with 
a dedicated relation called
$\texttt{comp}$ (\Cref{fig:tokenization}).
Note that $w_j$ has been removed.
As a result, in
$\mathcal{G}$ we have that: $\mathbf{PA}(W_{j_{1}}) = \mathbf{PA}(W_{j})$ and $\mathbf{PA}(W_{j_{2}}) = \mathbf{PA}(W_{j}) \cup \{ W_{j_{1}} \}$. 

\subsection{Experiments}
\label{subsec.experiments_appendix}

We use the  architectures described above  across all the experiments (e.g., the same number of blocks, learnable parameters, etc.). Specifically, we  freeze the weights of the visual encoder and we train only our textual decoder. We train in mixed precision (FP16) with batch size set to 128 on a GPU RTX A5000 with 24GB of VRAM for 10 epochs.
Following \citet{NegCLIP}, we select the best
 checkpoint using the  validation set provided in \citep{NegCLIP}. 
 In all the datasets and in all the experiments,  
we use the Adam optimizer with an initial learning rate set to $5\times10^{-4}$. Finally,  we apply a Cosine Annealing Learning Rate Scheduler with 50 warmup steps.

\section{Datasets and tasks}
\label{app.Datasets}

We provide details about the FG-OVD dataset and in \Cref{app.fg_ovd} and we briefly summarize here the main characteristics of the other benchmarks.
\Cref{tab:datasets_summary} shows the main statistics of each dataset and in \Cref{subsec.qualitative_results} we show a few images illustrating the benchmark typically tasks.

\begin{table}[ht]
\caption{Main statistics of the benchmarks.}
\label{tab:datasets_summary}
\centering
\resizebox{0.7\textwidth}{!}{%
\begin{tabular}{llrr}
\toprule
\textbf{Dataset} & \textbf{Split} & \textbf{Number of testing} & \textbf{Avg.  caption length} \\
 &  & \textbf{samples} & \textbf{(n. of words)} \\
\midrule
ARO                 & Relation & 23,937 & 8.1 \\
ARO                 & Attribution & 28,748 & 7.1 \\
\midrule
SugarCrepe          & Add & 2,754 & 12.9 \\
SugarCrepe          & Replace & 3,846 & 11.5 \\
SugarCrepe          & Swap & 911 & 13.5 \\
\midrule
VL-Checklist        & Attribute & 118,253 & 2.4 \\
VL-Checklist        & Object & 389,357 & 3.2 \\
VL-Checklist        & Relation & 75,641 & 3.5 \\
\midrule
ColorSwap           & - & 300 & 8.8 \\
\midrule
FG-OVD             & Hard & 3,545 & 10.4 \\
FG-OVD             & Medium & 2,968 & 11.3 \\
FG-OVD             & Easy & 1,299 & 16.2 \\
FG-OVD             & Trivial & 3,545 & 9.7 \\
\bottomrule
\end{tabular}}
\end{table}

{\bf ARO}
\citep{NegCLIP} is a VLM benchmark for compositional reasoning and word-order sensitivity. It is composed of two main tasks: Visual Genome Relation and Visual Genome Attribution. In the Visual Genome Relation task, the goal is to evaluate the models' ability to correctly interpret the relationships between objects. On the other hand, Visual Genome Attribution focuses on evaluating the ability to associate the correct attribute with the correct object. As mentioned in \Cref{sec.Experiments}, we do not use COCO Order and Flickr Order because different authors recently found that grammatical errors in the generated captions of these datasets  lead to tasks which can be solved purely relying on an LLM language prior \citep{zhang2024contrastingintramodalrankingcrossmodal,Cap,Revisiting}.

{\bf SugarCrepe}
\citep{sugarcrepe} is a dataset developed to evaluate how well VLMs can understand and process complex compositional tasks by presenting them with carefully designed hard negative examples. Drawing inspiration from datasets like CREPE \citep{crepe}, VL-CheckList \citep{vl_checklist}, and ARO \citep{NegCLIP}, SugarCrepe  focuses on atomic concepts and their compositions, such as objects, attributes, and relations. The dataset is split into three tasks: ``Replace'', ``Swap'' and ``Add''. In ``Replace'', an atomic concept in the original text is replaced with a new, mismatched concept. A replacement can involve an object, an attribute or a relation. In ``Swap'', the negative caption is created by exchanging two atomic concepts of the same category without introducing new elements. In ``Add'', a new concept is added to the original caption, leading to a misalignment  with the visual scene content.

{\bf VL-Checklist} 
\citep{vl_checklist} is a benchmark composed of four datasets: Visual Genome \citep{visual_genome}, SWiG \citep{swig}, VAW \citep{vaw}, and HAKE \citep{hake}. Each image is associated with two descriptions: a true and a false caption. The true descriptions originate from the original image-text pairs in the datasets, while the false ones are generated by modifying a single word in the true description, altering its overall meaning. These false descriptions are organized into three main types: objects, attributes, and relations.

{\bf ColorSwap} 
\citep{colorswap} evaluates the ability of multimodal models to accurately associate objects with their corresponding colors. Each sample contains a caption-image pair along with a ``color-swapped'' pair. The two captions in each sample use the same text, but the relation between colors and objects are inverted.

\subsection{Qualitative results}
\label{subsec.qualitative_results}

\begin{figure*}[!ht]
    \centering
   \includegraphics[width=\linewidth]{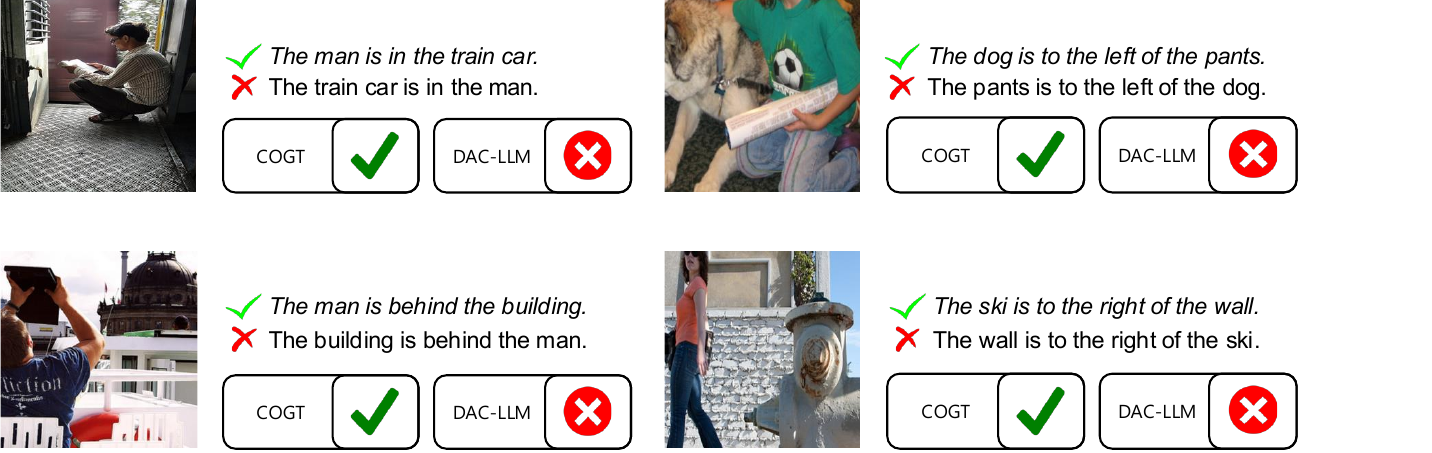}
    \caption{Qualitative results on sample images of the ARO Relation test split. We compare our approach with DAC-LLM which is the second best approach according to the results reported in \Cref{tab.SOTA-comparison}.}
    \label{fig.qual.1}
\end{figure*}

\begin{figure*}[!ht]
    \centering
   \includegraphics[width=\linewidth]{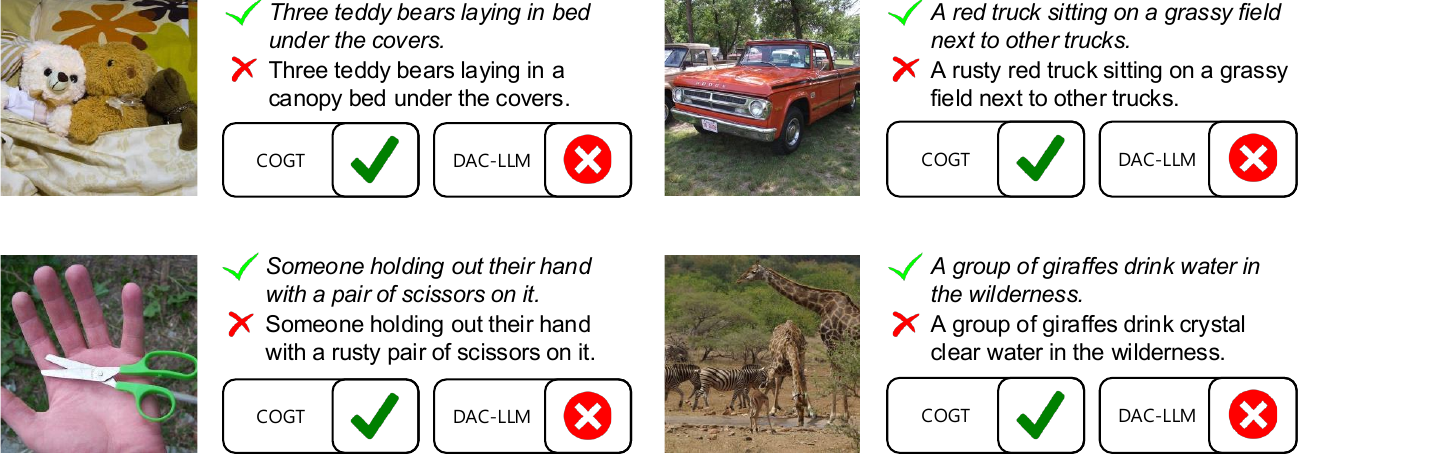}
    \caption{Qualitative results on sample images of SugarCrepe. }
    \label{fig.qual.2}
\end{figure*}

\begin{figure*}[!ht]
    \centering
   \includegraphics[width=\linewidth]{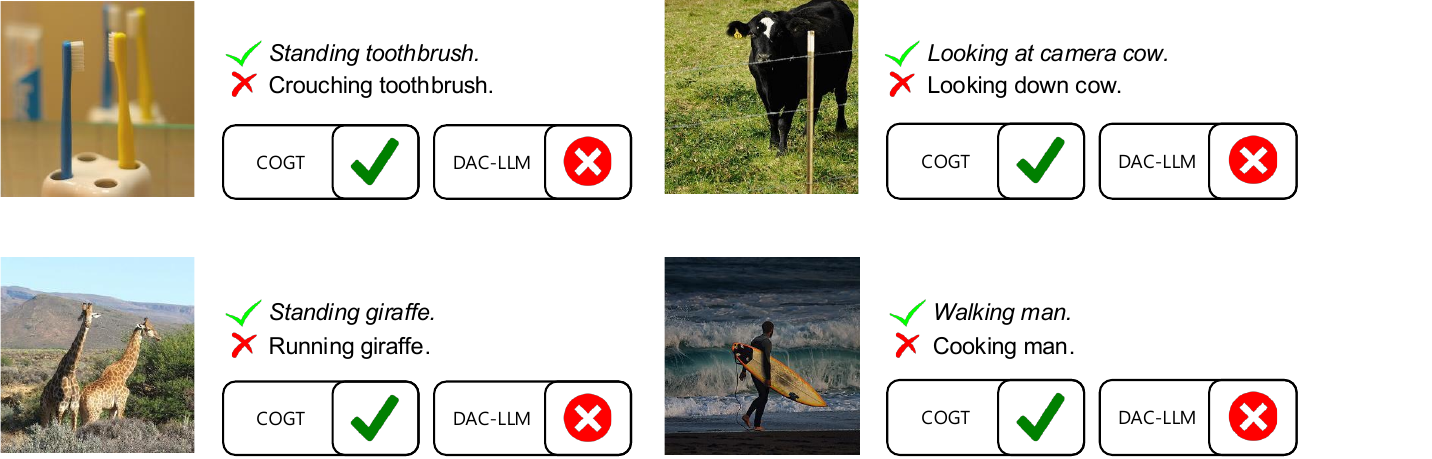}
    \caption{Qualitative results on sample images of VL-CheckList. }
    \label{fig.qual.3}
\end{figure*}

\begin{figure*}[!ht]
    \centering
   \includegraphics[width=\linewidth]{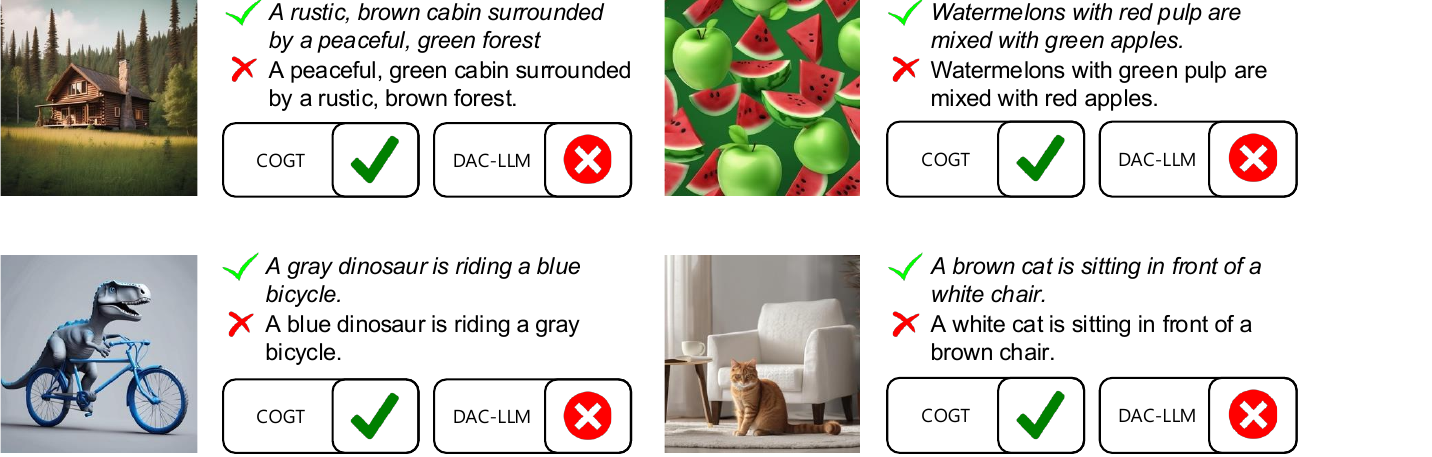}
    \caption{Qualitative results on sample images of ColorSwap. }
    \label{fig.qual.4}
\end{figure*}

\begin{figure*}[!ht]
    \centering
   \includegraphics[width=\linewidth]{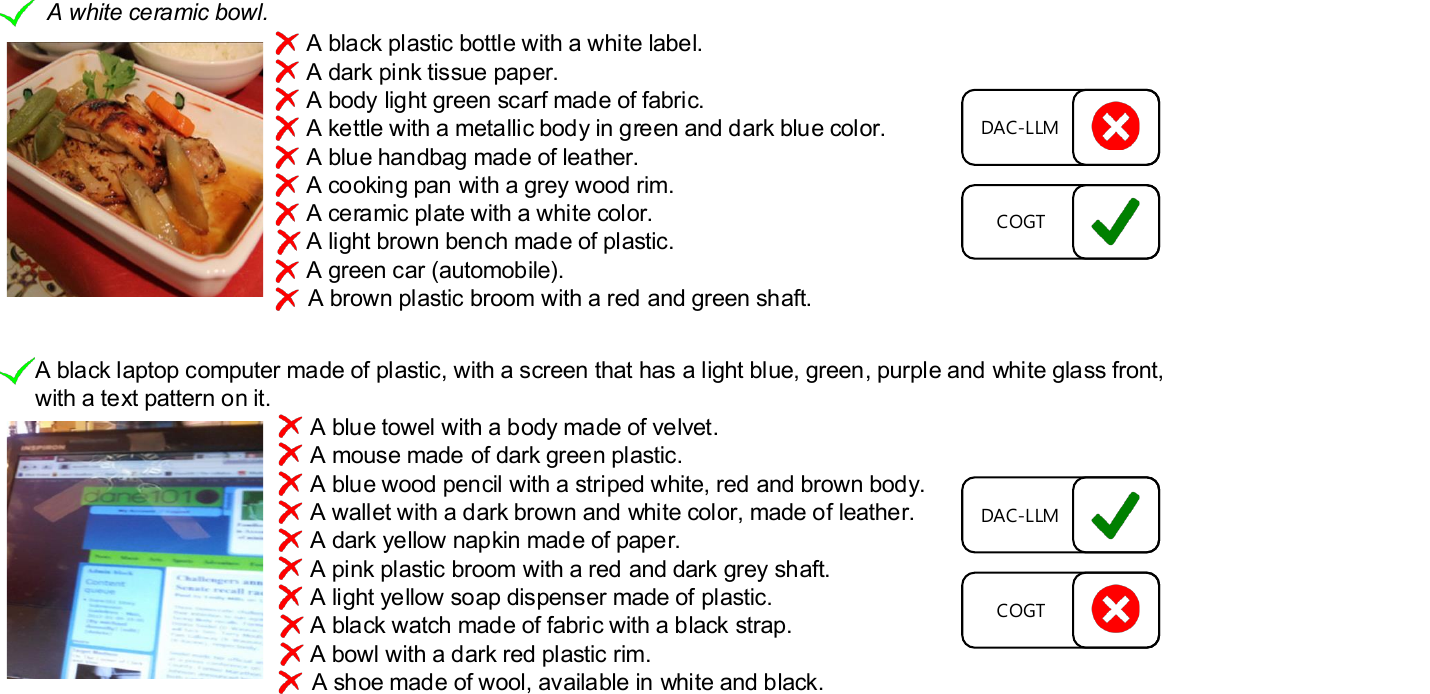}
   \caption{Qualitative results on sample images of the FG-OVD Trivial task. The image on the bottom shows a failure of  \ours.}
    \label{fig.qual.5}
\end{figure*}

\begin{figure*}[!ht]
    \centering
   \includegraphics[width=\linewidth]{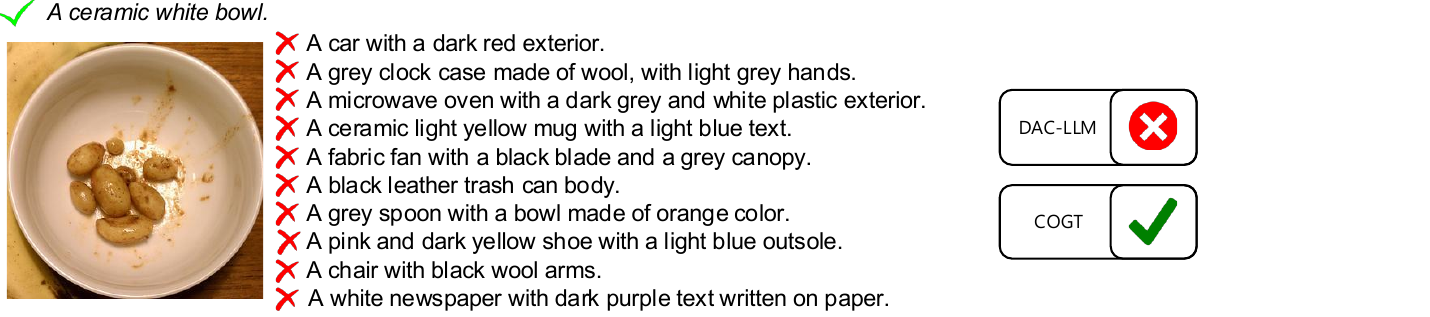}
   \caption{Qualitative results on sample images of the FG-OVD Trivial task where \ours is successful while DAC-LLM fails.}
    \label{fig.qual.6}
\end{figure*}

\begin{figure*}[!ht]
    \centering
    \includegraphics[width=\linewidth]{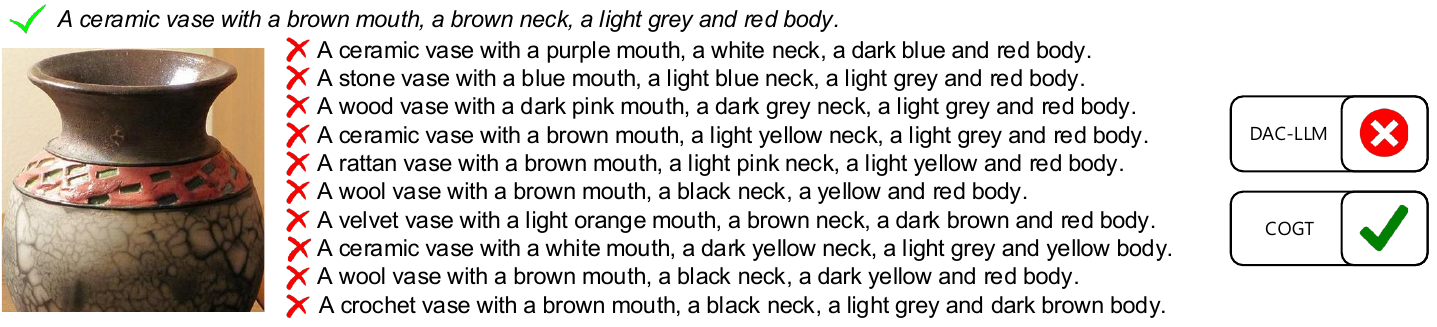}
   \caption{Qualitative results on sample images of the FG-OVD Easy task.}
    \label{fig.qual.7}
\end{figure*}

\begin{figure*}[!ht]
    \centering
    \includegraphics[width=\linewidth]{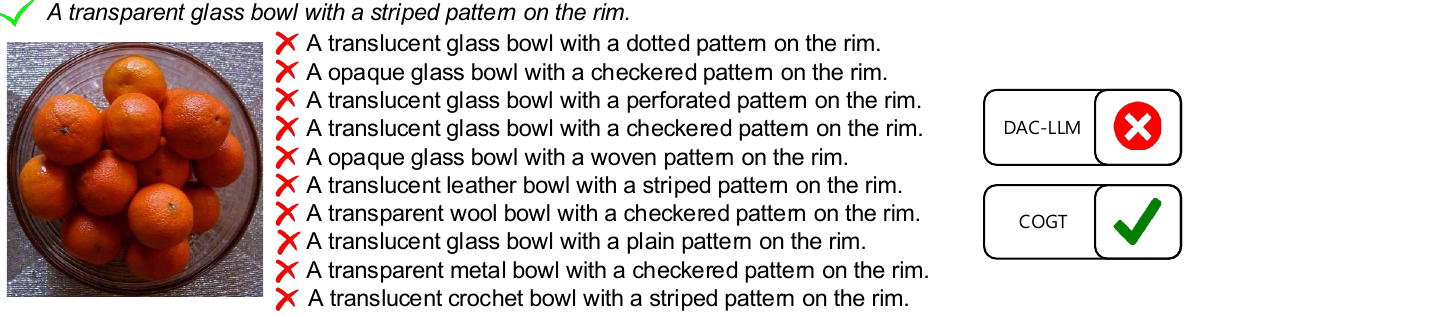}
   \caption{Qualitative results on sample images of the FG-OVD Medium task.}
    \label{fig.qual.8}
\end{figure*}

\begin{figure*}[!ht]
    \centering
    \includegraphics[width=\linewidth]{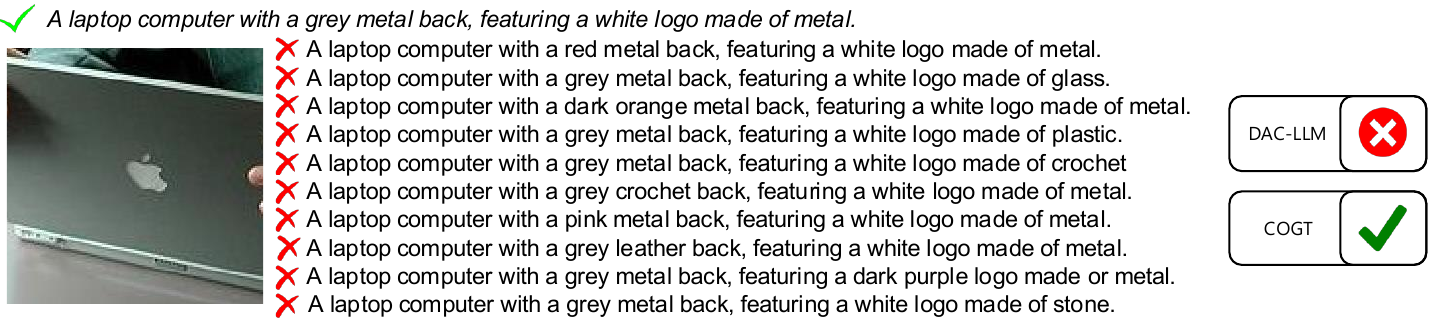}
   \caption{Qualitative results on sample images of the FG-OVD Hard task.}
    \label{fig.qual.9}
\end{figure*}

In \Cref{fig.qual.1}-\Cref{fig.qual.9} we show some qualitative results in which we compare \ours-CLIP+ with the second best approach in \Cref{tab.SOTA-comparison} (DAC-LLM).
We use these figures  also  to  illustrate the tasks of the different benchmarks, with a special emphasis on FG-OVD, proposed in this paper.

\clearpage
\section{Syntactic categories}
\label{subsec.syntactic_categories}

We report in \Cref{tab.syntactic_categories} the 45 syntactic categories defined in \citep{dependency_relations} and which form our set $V$ (\Cref{sec.Method}).
\begin{table}[h!]
\caption{List of the syntactic categories defined in \citep{dependency_relations}.}
\label{tab.syntactic_categories}
\centering
\begin{tabular}{@{}l@{}}
\toprule
\textbf{Syntactic Categories} \\ \midrule
\texttt{acomp} \\ 
\texttt{advcl} \\ 
\texttt{advmod} \\ 
\texttt{amod} \\ 
\texttt{appos} \\ 
\texttt{aux} \\ 
\texttt{auxpass} \\ 
\texttt{cc} \\ 
\texttt{ccomp} \\ 
\texttt{conj} \\ 
\texttt{cop} \\ 
\texttt{csubj} \\ 
\texttt{csubjpass} \\ 
\texttt{dep} \\ 
\texttt{det} \\ 
\texttt{discourse} \\ 
\texttt{dobj} \\ 
\texttt{expl} \\ 
\texttt{goeswith} \\ 
\texttt{iobj} \\ 
\texttt{mark} \\ 
\texttt{mwe} \\ 
\texttt{neg} \\ 
\texttt{nn} \\ 
\texttt{npadvmod} \\ 
\texttt{nsubj} \\ 
\texttt{nsubjpass} \\ 
\texttt{num} \\ 
\texttt{number} \\ 
\texttt{parataxis} \\ 
\texttt{pcomp} \\ 
\texttt{pobj} \\ 
\texttt{poss} \\ 
\texttt{possessive} \\ 
\texttt{preconj} \\ 
\texttt{predet} \\ 
\texttt{prep} \\ 
\texttt{prt} \\ 
\texttt{punct} \\ 
\texttt{quantmod} \\ 
\texttt{rcmod} \\ 
\texttt{root} \\ 
\texttt{tmod} \\ 
\texttt{xcomp} \\ \bottomrule
\end{tabular}
\end{table}

\end{document}